\documentclass[journal]{IEEEtran}
\usepackage{algorithm}
\usepackage{algpseudocode}
\usepackage{algorithmicx}
\usepackage{listings}
\usepackage{pifont}

\newcommand{\cmark}{\ding{51}}%
\newcommand{\xmark}{\ding{55}}%
\newcommand{\bmark}{\ding{61}}

\usepackage[utf8]{inputenc} 
\usepackage[T1]{fontenc}    
\usepackage{hyperref}       
\usepackage{url}            
\usepackage{booktabs}       
\usepackage{amsfonts}       
\usepackage{nicefrac}       
\usepackage{microtype}      
\usepackage{xcolor}         
\usepackage{threeparttable}
\usepackage{amsmath,amssymb,amsfonts}
\usepackage[pdftex]{graphicx}
\usepackage{multirow}
\usepackage{multicol}
\usepackage{adjustbox}
\usepackage{wrapfig}
\usepackage{booktabs}
\usepackage{rotating}
\newtheorem{theorem}{Theorem}

\def\mathbi#1{\textbf{\em #1}}
\ifCLASSINFOpdf
\else
\fi
\begin{document}
%
\title{Temporal Source Recovery for Time-Series Source-Free Unsupervised Domain Adaptation}
%
%
%

\author{Yucheng Wang,
        Peiliang Gong,
        Min Wu,
        Felix Ott,
        Xiaoli Li,~\IEEEmembership{Fellow, IEEE,}
        Lihua Xie,~\IEEEmembership{Fellow, IEEE,}
        and Zhenghua Chen
\thanks{Yucheng Wang is with Institute for Infocomm Research, A$^*$STAR, Singapore and the School of Electrical and Electronic Engineering, Nanyang Technological University, Singapore (Email: yucheng003@e.ntu.edu.sg).}
\thanks{Peiliang Gong is with College of Artificial Intelligence, Nanjing University of Aeronautics and Astronautics, China (Email: plgong@nuaa.edu.cn).}
\thanks{Felix Ott is with Fraunhofer Institute for Integrated Circuits, Germany (Email: felix.ott@iis.fraunhofer.de).}
\thanks{Min Wu and Zhenghua Chen are with Institute for Infocomm Research, A$^*$STAR, Singapore (Email: wumin@i2r.a-star.edu.sg, chen0832@e.ntu.edu.sg).}
\thanks{Xiaoli Li is with Institute for Infocomm Research, A$^*$STAR, Singapore and the College of Computing and Data Science, Nanyang Technological University, Singapore (Email: xlli@i2r.a-star.edu.sg).}
\thanks{Lihua Xie is with the School of Electrical and Electronic Engineering, Nanyang Technological University, Singapore (Email: elhxie@ntu.edu.sg).}}

%
%

\markboth{Journal of \LaTeX\ Class Files,~Vol.~14, No.~8, August~2015}%
{Shell \MakeLowercase{\textit{et al.}}: Bare Demo of IEEEtran.cls for IEEE Journals}
%



\maketitle

\begin{abstract}
Time-Series (TS) data has grown in importance with the rise of Internet of Things devices like sensors, but its labeling remains costly and complex. While Unsupervised Domain Adaptation (UDAs) offers an effective solution, growing data privacy concerns have led to the development of Source-Free UDA (SFUDAs), enabling model adaptation to target domains without accessing source data. Despite their potential, applying existing SFUDAs to TS data is challenging due to the difficulty of transferring temporal dependencies—an essential characteristic of TS data—particularly in the absence of source samples. Although prior works attempt to address this by specific source pretraining designs, such requirements are often impractical, as source data owners cannot be expected to adhere to particular pretraining schemes. To address this, we propose Temporal Source Recovery (TemSR), a framework that leverages the intrinsic properties of TS data to generate a source-like domain and recover source temporal dependencies. With this domain, TemSR enables dependency transfer to the target domain without accessing source data or relying on source-specific designs, thereby facilitating effective and practical TS-SFUDA. TemSR features a masking–recovery–optimization process to generate a source-like distribution with restored temporal dependencies. This distribution is further refined through local context-aware regularization to preserve local dependencies, and anchor-based recovery diversity maximization to promote distributional diversity. Together, these components enable effective temporal dependency recovery and facilitate transfer across domains using standard UDA techniques. Extensive experiments across multiple TS tasks demonstrate the effectiveness of TemSR, which even surpasses existing TS-SFUDA methods that require source-specific designs.
\end{abstract}


%
\IEEEpeerreviewmaketitle

\section{Introduction}
With the rapid growth of the Internet of Things, Time-Series (TS) data has been increasingly critical in multiple downstream tasks such as healthcare \cite{klepl2024graph,jin2024survey,ott2022domain} and industrial maintenance \cite{wang2024fully,chen2020machine}. While deep learning models yield promising results, they heavily depend on extensive labeled data, which is hard to obtain due to high labeling costs. 
To address this, Unsupervised Domain Adaptation (UDA) \cite{wilson2020survey,SEA++2024wang}, which transfers knowledge from a labeled source domain to an unlabeled target domain, has gained attention to reduce label reliance in TS tasks.

Although UDA techniques have proven effective, they typically require access to both source and target domains to bridge domain gaps. However, data privacy concerns prevent access to source data in many real-world scenarios \cite{li2024comprehensive}, leaving only a pretrained model available for adaptation. This challenge significantly limits the applicability of the standard UDA methods, as they are not designed for such restricted settings. To address this, Source-Free UDA (SFUDA) has been proposed, aiming to adapt the pretrained model to target domains without access to source data, and has shown promising results under such conditions. However, most existing techniques overlook the temporal dependencies critical to TS data \cite{MAPU2023ragab}, limiting their effectiveness for Time-Series SFUDA (TS-SFUDA).

TS data consist of continuous numerical measurements, where temporal dependencies are the most important characteristic, representing how temporal patterns evolve over time. For effective adaptation, transferring these dependencies across domains is crucial to learn domain-invariant features for TS data \cite{ragab2023adatime,Purushotham2017VariationalRA,dskn}. However, without accessing source data, directly transferring these dependencies becomes challenging. To address this, recent research \cite{MAPU2023ragab} has explored preserving temporal dependencies during source pretraining and restoring them during target adaptation. Although effective, these approaches require specific pretraining designs in the source domain, which are impractical for real-world applications. Thus, an effective and practical TS-SFUDA approach must meet two key criteria: \textit{1. Even without source data, the temporal dependencies can still be transferred across domains; 2. Additional designs during source pretraining should be avoided.}

Following this criterion, we aim to generate a source-like domain that resembles the original source distribution while recovering its temporal dependencies, allowing traditional UDA techniques to transfer such dependencies to the target domain. However, generating such a source-like TS domain poses two challenges. First, generating temporally meaningful samples requires careful handling of two inherent properties of TS data. On one hand, TS data normally exhibits temporal sparsity: individual time points contain sparse information, while meaningful temporal patterns typically emerge within contiguous patches. On the other hand, these temporal patches often exhibit strong correlations. Ignoring these sparsity and dependency properties hinders the recovery of temporal dependencies. Second, aligning the generated samples with the source domain is challenging. TS data is sensitive—even minor variations, such as slight amplitude changes, can cause significant shifts in temporal patterns and alter their meanings. As such, preserving fidelity to the source domain typically requires guidance from source properties—such as source distributions—to constrain the generation process. However, in the source-free setting, no direct supervisory signal is available for guidance, making it difficult for the generated samples to retain the temporal dependencies of the source domain.

To address these challenges, we propose Temporal Source Recovery (TemSR), which generates a source-like domain while recovering temporal dependencies based on TS properties, enabling effective transfer using traditional UDA techniques for TS-SFUDA. TemSR contains two steps: recovery and enhancement, which jointly restore and refine the dependency information. To recover temporal dependencies, it is essential to consider the patch dependency property, which captures correlations between patches and thus enables the inference of missing patches with unmasked contexts. Conversely, if a recovered sample—whose missing parts are completed based on contextual dependencies—aligns with the source domain, it serves as evidence that the underlying source temporal dependencies have been successfully restored. Based on this insight, we design a masking-and-recovery scheme: certain patches—comprised of time points—are masked in accordance with the sparsity property of TS data, and a recovery model is trained to infer the missing parts from unmasked parts.

\begin{figure}
    \centering
    \includegraphics[width=1\linewidth]{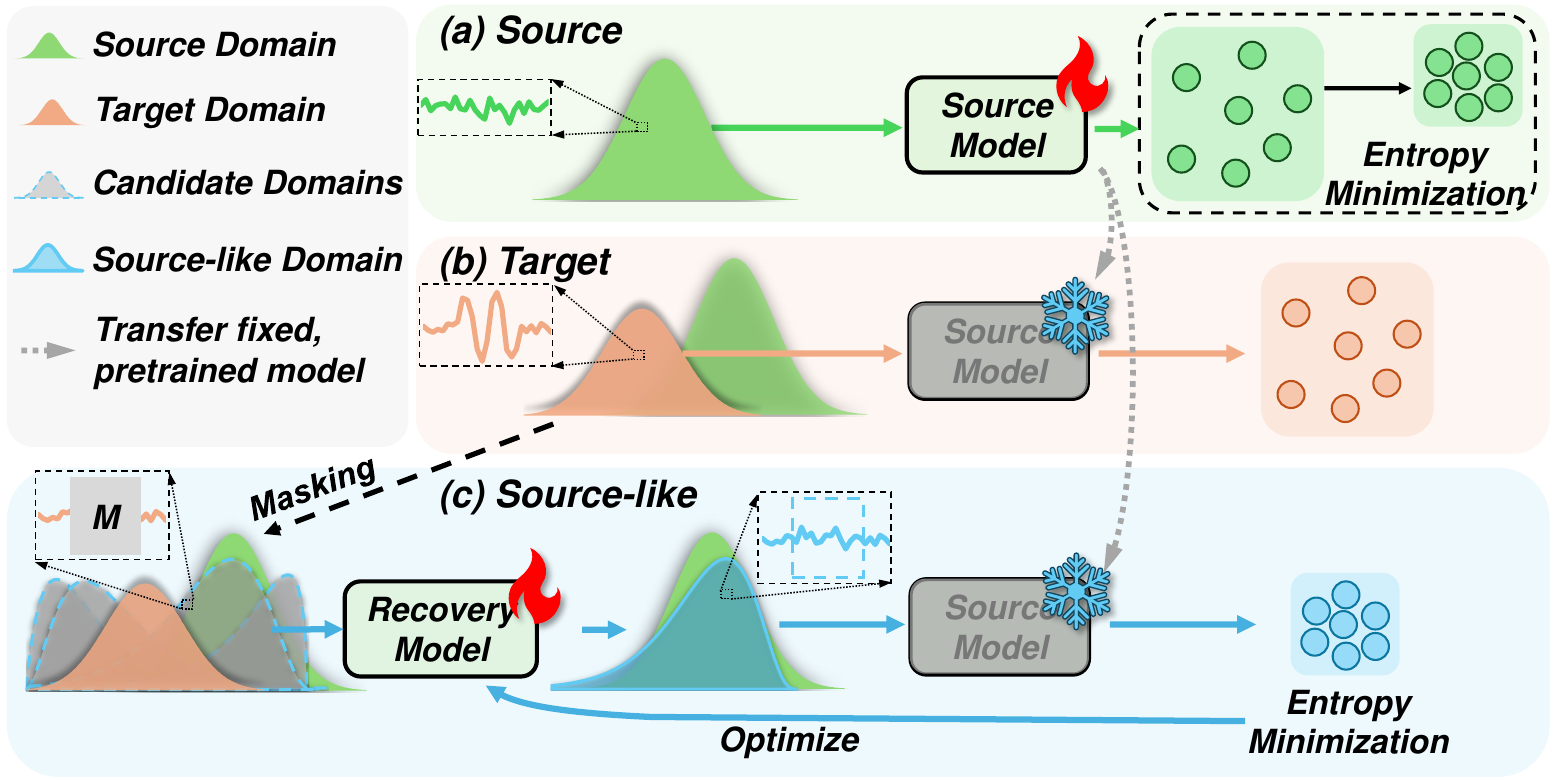}
    \caption{(a) The source model is trained via entropy minimization, which reduces prediction uncertainty and encourages deterministic outputs for domains with source characteristics; (b) Due to domain gaps, the fixed pretrained model yields high-entropy (uncertain) outputs on target data; (c) Masking introduces diversity into the target domain, implicitly providing candidate domains. To regain deterministic outputs, the input to the pretrained source model must exhibit source characteristics. Thus, entropy minimization inversely pushes the recovery model to generate source-like distributions.}
    \label{fig:intro_entropy}
    \vspace{-0.2cm}
\end{figure}

While potential, recovering source characteristics still requires aligning the recovered samples with the source domain, which is challenging in the source-free setting. To address this, we design an entropy-based optimization strategy that serves as indirect guidance from source properties. Specifically, we minimize the entropy of recovered samples, computed using the fixed pretrained source model. With minimized entropy on source data, the pretrained source model tends to produce deterministic outputs for distributions with source properties. By minimizing the entropy for recovered samples, this output constraint can inversely regularize these samples, forcing them to align with the source-like distribution, as shown in Fig. \ref{fig:intro_entropy}. Notably, this optimization focuses on sample-level recovery for global temporal dependencies, potentially overlooking local temporal dependencies—short-scale contextual information essential for preserving natural temporal structure. These local dependencies capture consistent transitions of patterns, which are crucial for modeling effective temporal dynamics. For example, in human activity recognition using wearable sensors, running involves rhythmic transitions between foot-strike and toe-off phases, which provide key cues for identifying the activity. Failing to recover such patterns degrades temporal coherence, leading to suboptimal recovery. To solve this, we enhance optimization as local context-aware regularization by minimizing the entropy of local patch sets extracted from recovered samples, thus better restoring the temporal structure.

Effective recovery also depends critically on the masking ratio, which introduces the diversity necessary to effectively recover a source-like distribution. However, this presents a challenge: while a high masking ratio increases diversity, it risks causing the recovery model to collapse into producing trivial outputs to minimize entropy. Conversely, a low masking ratio mitigates this issue but lacks sufficient diversity, hindering effective recovery of the source-like distribution. To enhance the recovery, we introduce an enhancement module, anchor-based recovery diversity maximization, where recovery diversity maximization enhances diversity in recovered samples and anchors ensure this diversity aligns with the source distribution. By effectively enhancing diversity, this module facilitates the recovery of an optimal source-like distribution.

Our contributions are threefold. 1. We design a recovery process involving masking, recovery, and optimization to generate a source-like distribution with recovered source temporal dependencies, which is further refined by local context-aware regularization to improve local temporal recovery. 2. We design an enhancement module to improve diversity in the source-like distribution through anchor-based recovery diversity maximization, with anchors ensuring this diversity aligns with the source distribution. By effectively enhancing diversity, this module facilitates the recovery of an optimal source-like distribution. 3. Extensive experiments across various TS tasks indicate the effectiveness of TemSR, which even surpasses the TS-SFUDA method that requires source pretraining designs. Additional analysis on distribution discrepancy changes between source, source-like, and target domains further verifies TemSR's ability to recover an effective source-like domain and reduce gaps between the source and target domains even without accessing the original source data.

\begin{figure*}[ht!]
    \centering
    \includegraphics[width=.93\linewidth]{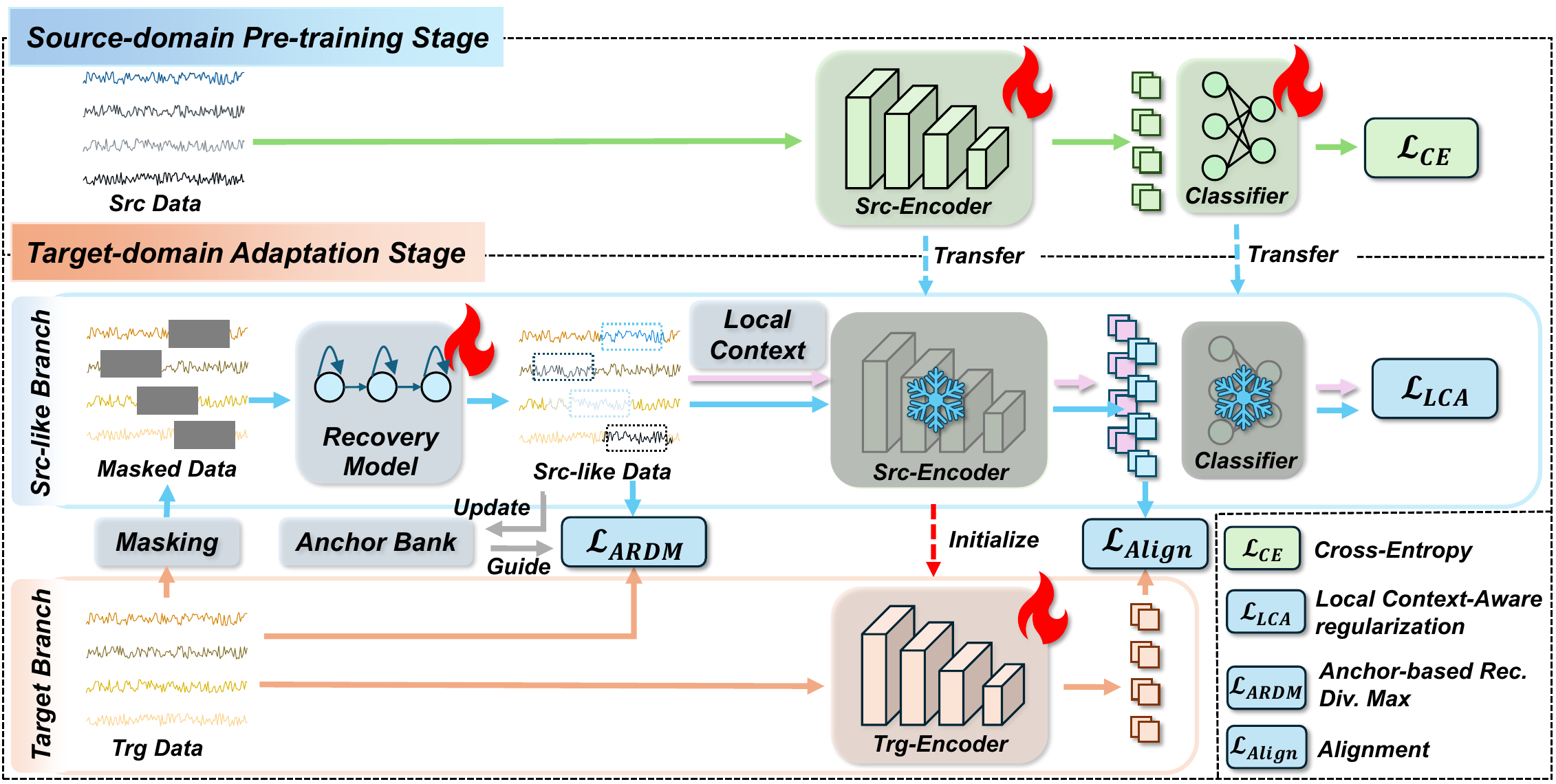}
    \caption{Overall TemSR. An encoder pretrained on a source domain is transferred to a target domain for adaptation without source data, using source-like and target branches. In the source-like branch, masked target samples are recovered by a recovery model. With the fixed source encoder, their entropy is computed via a local context-aware regularization loss $\mathcal{L}_{LCA}$ and minimized for optimization to generate a source-like distribution with restored temporal dependencies. Meanwhile, an Anchor-based Recovery Diversity Maximization loss $\mathcal{L}_{ARDM}$ enhances the diversity of the generated distribution for effective recovery. Finally, source-like and target distributions are aligned with an alignment loss $\mathcal{L}_{Align}$, enabling the transfer of temporal dependencies.}
    \vspace{-0.3cm}
    \label{fig:overall}
\end{figure*}
\section{Related Work}
\subsection{Time-Series Unsupervised Domain Adaptation}
To reduce label reliance in TS tasks, UDA methods have been widely applied by transferring knowledge from a labeled source domain to an unlabeled target domain. A key challenge in TS UDA is transferring temporal dependencies across domains to learn domain-invariant features \cite{ragab2023adatime}. To address this, existing methods primarily adopt metric-based or adversarial-based strategies. Metric-based approaches capture temporal structures by extracting sequence-level features and aligning their statistical distributions across domains \cite{dskn, raincoat, SASA2021chen}. In contrast, adversarial-based methods utilize domain discriminators to enforce temporal consistency and promote domain-invariant representations \cite{wilson2020multi, wilson2023calda, Purushotham2017VariationalRA}. More recently, contrastive learning has also been explored to enhance temporal feature robustness in TS UDA \cite{eldele2023contrastive, ozyurt2022contrastive}. In parallel, studies have investigated channel dependencies in TS data to further improve generalization abilities \cite{SEA2022wang, SEA++2024wang}. Despite their effectiveness, most existing TS UDA approaches require access to source data, which may be impractical due to privacy or data-sharing constraints. This highlights the need for SFUDA, where adaptation is performed without access to the source data.

\subsection{Source-Free Unsupervised Domain Adaptation}
To enable effective UDA without accessing source data, researchers have explored SFUDA through two primary paradigms: model-based and data-based methods \cite{fang2024source}. Model-based approaches focus on adapting a source-pretrained model to the target domain through various self-supervised techniques, including entropy minimization that encourages confident predictions \cite{Entropymin1, Entropymin2}, pseudo-label generation methods that iteratively refine target labels \cite{nrc, Pseudolabel1, Pseudolabel2}, and contrastive learning frameworks that align feature representations \cite{DAC, Contrast1}. These techniques modify the model's behavior to better suit the target domain while preserving knowledge from the source. Data-based methods take a fundamentally different approach by attempting to generate source domain characteristics from target data. This includes selective sampling techniques that identify target instances resembling source distributions \cite{Select1, Select2}, as well as generative adversarial networks (GANs) that synthesize source-like samples from target data \cite{GAN2}. The synthetic source distributions enable the application of conventional UDA techniques while maintaining source-free constraints.

Although existing methods perform well, particularly in visual tasks, they are not well suited for TS data as they fail to consider the core TS property. Unlike images, TS data lacks universal structures—a signal shape (e.g., a spike) can hold different meanings across domains. Without transferable temporal dependencies to align these meanings, model-based methods struggle to adapt pretrained models to learn useful patterns. Data-based approaches suffer the same issue: their performance hinges on the quality of generated distributions. Without explicitly modeling temporal relationships, these generated samples fail to preserve critical temporal coherence, degrading performance in TS-SFUDA. While video data also exhibits temporal dynamics \cite{sahoo2021contrast,wei2023unsupervised}, video-based methods cannot be directly applied to TS due to key differences: Unlike video data composed of dense image frames, TS is sparse and low-dimensional at each time point, lacking the rich spatial semantics that aid pattern recovery. Video models benefit from strong spatial context and visual priors (e.g., edges, textures), making temporal coherence a secondary cue. In contrast, TS lacks such priors and relies entirely on temporal dependencies embedded in abstract numerical signals, making transferable representation learning more challenging. Moreover, video dynamics are typically stable across domains (e.g., walking appears similarly in different videos), while TS dynamics are sensitive—small variations can drastically change meaning, complicating cross-domain alignment. These fundamental differences highlight the need for frameworks that can model and transfer TS-specific temporal structures effectively \cite{ozyurt2022contrastive, wang2024pond, dskn, SEA++2024wang, wilson2023calda, liu2024boosting}.


Currently, research in TS-SFUDA remains limited \cite{MAPU2023ragab, furqon2025time}, as the absence of source data poses significant challenges in transferring temporal dependencies to the target domain. Ragab et al. \cite{MAPU2023ragab} addressed this by introducing a temporal imputer during the source pretraining phase to store temporal dependencies, which are then restored in the adaptation stage using the fixed imputer. While this approach demonstrates that temporal dependencies can be transferred under the TS-SFUDA setting, it relies on additional source-specific pretraining steps. This is impractical in many real-world scenarios, as source data holders cannot be expected to follow customized pretraining procedures. To address this, we propose TemSR, a framework that effectively transfers temporal dependencies across domains without requiring any additional operations during source pretraining, thereby ensuring both practical deployment and strong performance in the TS-SFUDA setting.

\section{Methodology}

\subsection{Problem Definition}
Given a labeled source domain $\mathbb{D}_S = \{\mathbi{X}_S^i, {y}_S^i\}_{i=1}^{n_S} $ with $n_S$ samples and an unlabeled target domain $\mathbb{D}_T= \{\mathbi{X}_T^i\}_{i=1}^{n_T} $ with $n_T$ samples, $\mathbi{X}^i_{\{S,T\}}\in\mathbb{R}^{N\times{L}}$ represents TS data consisting of $L$ temporally correlated time points, each collected from $N$ channels, and $y_S^i$ denotes the corresponding source label. In the TS-SFUDA setting, an encoder $\mathcal{F}_{\theta}$ and a classifier $\mathcal{G}_{\phi}$ are pretrained on the labeled source domain. After pretraining, the pretrained encoder is adapted to the unlabeled target domain without access to source data. 

Given the critical role of temporal dependencies in TS data, transferring the dependencies across domains is essential for successful TS-SFUDA. However, in the absence of source data, it becomes difficult to directly observe or characterize the source domain’s temporal dynamics, such as correlations between patches or recurrent structures in samples. Moreover, the parameters in the pretrained source model do not explicitly encode these fine-grained temporal structures. These limitations hinder the transfer of temporal dependencies to the target domain. To address this, we propose generating a source-like domain that approximates the original source distribution and recovers source temporal dependencies. This recovered domain enables the use of conventional UDA techniques to transfer the dependency information, thereby facilitating the adaptation of the pretrained encoder to the target domain. Once the encoder is adapted, its extracted features can be passed to the pretrained classifier for downstream prediction tasks.
\subsection{Overall Framework}
\label{sec:overall_framework}
Fig. \ref{fig:overall} presents the overall TemSR. After an encoder is pretrained on the labeled source domain, TemSR adapts it to the unlabeled target domain without access to source data. The framework consists of two branches: a source-like branch and a target branch. The source-like branch aims to generate a source-like distribution with recovered temporal dependencies. Specifically, target samples are masked and then completed by a recovery model. The recovered samples are subsequently passed through the fixed, pretrained source encoder to compute their entropy losses by local context-aware regularization. The resulting entropy loss is minimized to guide the recovery model in generating source-like samples with restored temporal structure. To further enhance recovery, we introduce an anchor-based recovery diversity maximization loss that encourages diversity among recovered samples while ensuring this diversity remains aligned with the source domain through anchors, whose quality is improved by an anchor bank. Finally, a distribution alignment loss aligns the source-like and target domains, facilitating the transfer of temporal dependencies and thus improving the adaptation of the encoder—initialized from the source model—to the target domain. Further details are provided in the following sections.

\subsection{Recovery}
The recovery process begins with an initialized distribution, where masking is applied to introduce diversity into the initialized samples. These masked samples are then completed by a recovery model and optimized through the local context-aware regularization, yielding a source-like distribution with recovered source temporal dependencies.
\paragraph{Initialization}
\begin{figure}[htbp]
    \centering
    \includegraphics[width=0.72\linewidth]{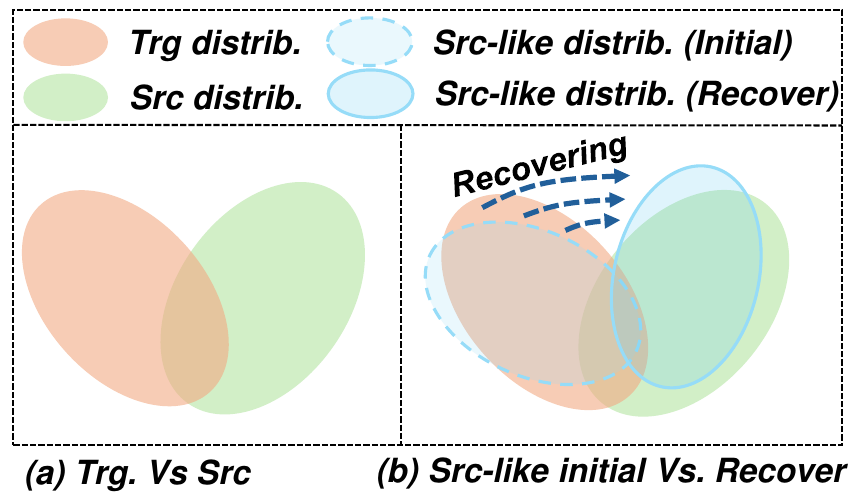}
    \caption{(a) Source and target distributions are distinct but related. (b) Source-like distribution, when initialized from the target distribution, can more easily be optimized to resemble source distribution.}
    \label{fig:src_initial}
\end{figure}

A critical step in generating an effective source-like distribution is proper initialization, for which we identify two key requirements:

    1. The initialized distribution should be close to the source distribution; otherwise, obtaining an effective source-like distribution is difficult.
    
    2. The time points of the initialized samples must be continuous, as random time points would hinder the recovery of source temporal dependencies.

Existing generative methods, such as GANs \cite{yang2023casting,GAN2}, fail to satisfy these requirements for two key reasons. First, they typically initialize the generator with a random noise vector sampled from a standard distribution (e.g., Gaussian or uniform). This stochastic initialization often diverges substantially from the source distribution, expanding the solution space and hindering convergence to an optimal source-like distribution. Second, they lack inherent mechanisms to model sequential data or temporal dependencies, as they generate each sample independently without enforcing inter-point continuity. Consequently, they may produce temporally incoherent time points, failing to capture the underlying dynamics of the data.

To solve this, initializing the source distribution using the target distribution offers an effective solution. As UDA typically operates on different but related domains, the target distribution is normally not significantly different from the source distribution, as shown in Fig. \ref{fig:src_initial}. By initializing a source-like domain with the target domain, we can simplify the optimization process and also preserve the continuity of time points in the samples.

\paragraph{Masking and Recovery}
\label{sec:masking_recovery}
With the initialized distribution, we introduce diversity to allow optimization toward a source-like distribution. Masking is an effective approach, as it not only introduces diversity but also aids in recovering temporal dependencies. By masking portions of TS data, a recovery model is forced to infer masked portions using contextual information from the unmasked parts. Successfully recovering these segments requires the model to understand how temporal patterns evolve, thereby encouraging it to capture the underlying temporal dependencies. Thus, if the recovered sample—whose missing parts are completed based on unmasked contexts—aligns with the source domain, it provides evidence that the source temporal dependencies have been successfully recovered. Based on this insight, as shown in Fig. \ref{fig:overall}, specific patches of TS sequences, consisting of time points, are masked in accordance with the sparsity property of TS data, with the masked portions determined by a masking ratio $p_m$. Given a target sample $\mathbi{X}^i_T$, masking generates its masked form $\Bar{\mathbi{X}}^i_T$= $M(\mathbi{X}^i_T)$, which is then completed by a recovery model $\mathcal{R}_{\zeta}$ to generate $\mathbi{X}^i_{Sl} = \mathcal{R}_{\zeta}(M(\mathbi{X}^i_T))$. These recovered samples are then optimized to align with the source domain.

\paragraph{Optimization}
\label{sec:optimization}
Recovering source characteristics requires aligning the recovered samples with the source domain. However, this is inherently challenging in the source-free setting, where the source domain is inaccessible. To solve this, we propose leveraging the pretrained source model with entropy minimization as indirect guidance. Entropy minimization is widely used in model adaptation, as models with minimized entropy can produce deterministic outputs, and this ideal output constraint can inversely guide adaptation \cite{li2024comprehensive,liang2020we}. Inspired by this, we introduce entropy minimization to optimize the recovered samples. With the minimized entropy on source data, the pretrained model can produce deterministic outputs for distributions with source characteristics, as shown in Fig. \ref{fig:intro_entropy} (a). To retain such deterministic outputs on the recovered domain, the recovered samples—when passed through the fixed pretrained source model—must also exhibit source characteristics. Consequently, minimizing their entropy, as computed by the fixed model, can inversely force the recovered domain to align with the source distribution, as shown in Fig. \ref{fig:intro_entropy} (c). Additionally, the recovery model is forced to capture source temporal dependencies, as only by understanding the dependencies can the model effectively infer masked parts, minimize entropy, and finally ensure recovered samples to align with the source distribution.

While the recovery process can generate a source-like distribution, it primarily targets sample-level recovery, focusing on global temporal dependencies by capturing long-range patterns across entire sequences. However, this overlooks local temporal dependencies—short range correlations among patches that encode short-scale contextual information essential for preserving natural temporal structure. These local dependencies capture consistent transitions or repetitive sub-patterns that are critical for modeling temporal dynamics. For example, in human activity recognition, running involves recurring local patterns, such as rhythmic transitions between foot-strike and toe-off phases. These short-scale patterns provide discriminative cues for identifying the activity. If such local dependencies are not properly recovered, the temporal coherence of these patterns may degrade, leading to suboptimal recovery of temporal dependencies. To address this, we enhance the optimization as local context-aware regularization, which further minimizes the entropy of local patch sets extracted from recovered samples. This encourages local contexts—capturing short-range dependencies—to exhibit low entropy and align with the source distribution, thereby improving temporal structure recovery. Specifically, three types of local contexts, each corresponding to a particular patch set extracted from the recovered sample $\mathbi{X}^i_{Sl}$ using an extraction proportion $p_s$, are defined to capture and restore local temporal dependencies across different regions (see Appendix for details).

1. Early Context $\mathbi{X}^i_{Sl, E}$: Extracts the first $p_s$ proportion of $\mathbi{X}^i_{Sl}$ as a patch set, capturing local dependencies in the early temporal stage.

2. Late Context $\mathbi{X}^i_{Sl, L}$: Extracts the last $p_s$ proportion of $\mathbi{X}^i_{Sl}$ as a patch set, capturing local dependencies in the later temporal stage.

3. Recovered Context $\mathbi{X}^i_{Sl, R}$: Extracts all recovered (previously masked) portions as a patch set, ensuring they have minimized entropy and align with the source-like distribution.

Along with the complete version $\mathbi{X}^i_{Sl}$ representing the global context for sample-level recovery, denoted as $\mathbi{X}^i_{Sl, C}$ for consistency, we compute the entropy as Eq. (\ref{eq:LCAent}), where $\mathcal{F}_{\theta}(\cdot)$ represents the fixed pretrained source model, $\mathbb{S} = \{C, E, L, R\}$ denotes the set comprising the complete sample and various local contexts. By minimizing their entropy, we encourage both global and local contexts to align with the source distribution, thereby restoring global and local temporal dependencies and enhancing overall temporal structure recovery.
\begin{equation}
\label{eq:LCAent}
    \mathcal{L}_{LCAE} = -\sum_{k \in \mathbb{S}} \sum_{i} \mathcal{G}_{\phi}(\mathcal{F}_{\theta}(\mathbi{X}^i_{Sl,k})) \log \mathcal{G}_{\phi}(\mathcal{F}_{\theta}(\mathbi{X}^i_{Sl,k})).
\end{equation}
In addition to minimizing entropy, maintaining similar entropy across different contexts is also essential. Significant discrepancies in entropy may indicate disruptions in the flow of temporal information, implying that the model fails to capture smooth dependencies within the recovered sequences. To address this, we enforce entropy consistency across different contexts, as defined in Eq. (\ref{eq:siment}). This consistency constraint helps TemSR preserve a uniform temporal structure throughout the sequence.
\begin{align}
\label{eq:siment}
    \mathcal{L}_{LCAS} = \sum_{(k, s) \in \mathbb{S}}  &\left( \sum_i \mathcal{G}_{\phi}(\mathcal{F}_{\theta}(\mathbi{X}^i_{Sl,k} )) \log \mathcal{G}_{\phi}(\mathcal{F}_{\theta}(\mathbi{X}^i_{Sl,k} ) ) \right. \nonumber \\
    & \left. - \sum_i \mathcal{G}_{\phi}(\mathcal{F}_{\theta}(\mathbi{X}^i_{Sl,s} ) ) \log \mathcal{G}_{\phi}(\mathcal{F}_{\theta}(\mathbi{X}^i_{Sl,s} ) ) \right).
\end{align}
By combining the two losses, we define the local context-aware regularization loss as $\mathcal{L}_{LCA} = \mathcal{L}_{LCAE} + \mathcal{L}_{LCAS}$. By minimizing $\mathcal{L}_{LCA}$, we effectively generate a source-like distribution with recovered source temporal dependencies.

\subsection{Enhancement}
\label{sec:enhancement}
To optimize the initial distribution as a source-like distribution, masking introduces the essential diversity required for effective recovery. However, masking presents challenges. A large masking ratio can introduce sufficient diversity, increasing the chances of finding an optimal solution. However, it risks model collapse, where the recovery model shortcuts the learning process by filling masked parts with constant values, minimizing entropy without capturing the true underlying structure, as proof in Section \ref{sec:trivial_solution}. Using a small masking ratio avoids this collapse but fails to provide enough diversity for the model to learn an optimal source-like distribution. 

\begin{figure}[htbp]
    \centering
    \includegraphics[width=0.85\linewidth]{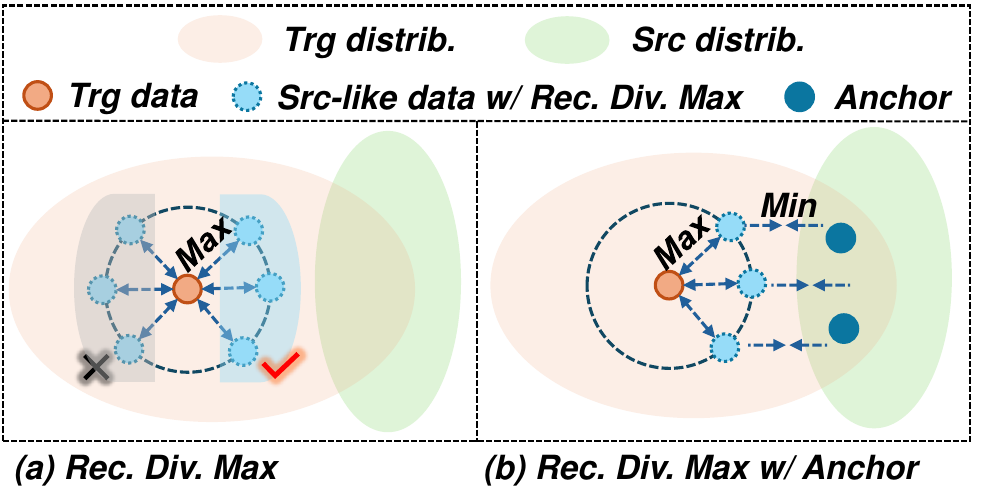}
    \caption{(a) Recovery diversity maximization causes recovered samples to deviate in unintended directions without proper constraints. (b) Anchors act as reference points, balancing diversity with fidelity to the source domain.}
    \label{fig:ARDM}
\end{figure}

\paragraph{Anchor-based Recovery Diversity Maximization} To effectively enhance diversity for optimal recovery, we introduce the anchor-based recovery diversity maximization module. This module encourages recovery diversity by maximizing the distance between recovered samples and their original samples. By pushing the recovered samples to diverge from their original forms, the samples are forced to enhance diversity (proof in Section \ref{sec:diversity}), allowing to explore a broader range of features that are crucial for capturing the complexity of the source distribution. However, without proper constraints, this recovery diversity maximization may cause the recovered samples to deviate in unintended directions, as shown in Fig. \ref{fig:ARDM} (a), leading to distributions that are not aligned with the source domain and hurting performance. To prevent this, we further introduce anchors to guide the process and ensure that the diversity remains consistent with the source distribution. Anchors act as reference points as shown in Fig. \ref{fig:ARDM} (b), balancing diversity with fidelity to the source domain.

\paragraph{Anchor Generation with Anchor Bank} 
\label{sec:Anchor_Generation_with_Anchor_Bank}
To effectively guide optimization toward the source distribution, generating high-quality anchors is crucial, as poor anchors can mislead the model and degrade performance. For optimal guidance, these anchors must closely align with the original source distribution. Thus, we propose selecting recovered samples with the lowest entropy, as they are more likely to reflect the source distribution and serve as ideal guides for the recovery process. 

While a simple approach is to select low-entropy samples from each batch, this may miss optimal candidates due to batch randomness. To address this, we implement an anchor bank $\mathbb{A}$, inspired by \cite{wu2018unsupervised}, to store all recovered samples with their corresponding entropy values:
\begin{equation}
\begin{split}
    \label{eq:anchor_bank}
    &\mathbb{A} = \{\mathbi{X}_{Sl}^i, H(\mathbi{X}_{Sl}^i)\}_{i=1}^{n_T},\\
    &H(\mathbi{X}_{Sl}^i) = -\mathcal{G}_{\phi}(\mathcal{F}_{\theta}(\mathbi{X}^i_{Sl})) \log \mathcal{G}_{\phi}(\mathcal{F}_{\theta}(\mathbi{X}^i_{Sl})).
\end{split}
\end{equation}
To ensure its quality, the anchor bank $\mathbb{A}$ is continuously updated during adaptation (as shown in Fig. \ref{fig:overall}) using a first-in-first-out strategy: the oldest samples are popped from the bank as new samples are pushed in.

From the anchor bank, we extract $k$ samples with the lowest entropy and compute their average to generate a representative anchor as shown in Eq. (\ref{eq:anchor_gene}), where $k$ is set by an anchor ratio.
\begin{align}
\label{eq:anchor_gene}
\mathbb{A}_k &= \left\{ \mathbf{X}_{Sl}^{j} \,\middle|\, j \in \operatorname*{arg\,min}^{(k)}_{i \in \{1,\dots,n_T\}} H(\mathbf{X}_{Sl}^i) \right\}, \\
\bar{\mathbf{X}}_A &= \frac{1}{k} \sum_{\mathbf{X}_{Sl}^j \in \mathbb{A}_k} \mathbf{X}_{Sl}^j.
\end{align}

\paragraph{Objective Loss} We have two key objectives: 1. Recovery Diversity Maximization: Maximize the distances between the recovered samples and their original samples; 2. Anchor Guidance: Minimize the distances between the recovered samples and the anchor sample. However, directly pushing all recovered samples toward the anchor risks collapse, where diversity is lost as all samples converge to a single point. To prevent this, we introduce an additional objective: 3. Anchor Guidance Auxiliary: Maximize the distances between any two recovered samples, ensuring variations among them. To achieve these three objectives simultaneously, the InfoNCE loss for contrastive learning is adopted \cite{eldele2021time}, which pulls the recovered samples toward the anchor while pushing them apart from each other and their original forms. Particularly, given recovered source-like samples $\mathbi{X}_{Sl}^i$, original target samples $\mathbi{X}_T^i$, and the anchor $\bar{\mathbi{X}}_A$, the anchor-based recovery diversity maximization loss is defined as Eq. (\ref{eq:ARDM}), where $B$ is batch size.
\begin{equation}
    \label{eq:ARDM}
    \begin{split}
        &\mathcal{L}_{ARDM} = - \frac{1}{B} \sum_{i=1}^{B} \log \frac{\mathcal{S}(\mathbi{X}_{Sl}^i, \bar{\mathbi{X}}_A)}{\mathcal{S}(\mathbi{X}_{Sl}^i, \mathbi{X}_T^i) + \sum_{k \neq i} \mathcal{S}(\mathbi{X}_{Sl}^i, \mathbi{X}_{Sl}^k)},\\
    &\mathcal{S}(\mathbi{X}_{Sl}^i,\mathbi{X}_*) = \exp\left( \mathcal{F}_{\theta}(\mathbi{X}_{Sl}^i)(\mathcal{F}_{\theta}(\mathbi{X}_*))^T / \tau \right).
    \end{split}
\end{equation}

\subsection{Adaptation}
Once the source-like distribution with source temporal dependencies is generated, we transfer this information to the target domain for adaptation. As the dependency information has already been recovered, traditional UDA techniques, such as metric-based methods, can be utilized for effective transfer \cite{wilson2023calda,Purushotham2017VariationalRA,SEA++2024wang}. For adaptation, we fine-tune the target encoder $\Bar{\mathcal{F}}_{\Bar{\theta}}$, initialized from the pretrained source encoder $\mathcal{F}_{\theta}$, to adapt to the target domain, as shown in Fig. \ref{fig:overall}. To further preserve target domain information, we incorporate target entropy minimization following \cite{liang2020we}, i.e., $\mathcal{L}_{TrgE} = - \sum_{i} \mathcal{G}_{\phi}(\Bar{\mathcal{F}}_{\Bar{\theta}}(\mathbi{X}^i_T)) \log \mathcal{G}_{\phi}(\Bar{\mathcal{F}}_{\Bar{\theta}}(\mathbi{X}^i_T))$. Eq. (\ref{eq:final_loss}) shows the final loss, where $\mathcal{L}_{Align}$ denotes the alignment loss computed by following the previous work \cite{SEA++2024wang}. Additionally, as $\mathcal{L}_{LCA}$ and $\mathcal{L}_{ARDM}$ determine the quality of the recovered source-like domain, the coefficients $\lambda_{LCA}$ and $\lambda_{ARDM}$ are introduced to appropriately adjust their contributions for optimal recovery.
\begin{equation}
    \label{eq:final_loss}
    \min \mathcal{L} = \lambda_{LCA}\mathcal{L}_{LCA} + \lambda_{ARDM}\mathcal{L}_{ARDM} + \mathcal{L}_{Align} + \mathcal{L}_{TrgE}.
\end{equation}
Notably, the source-like distribution may have poor quality during initial epochs, so adaptation at this stage could cause negative transfer. To solve this, we divide the adaptation process into source-like optimization and transfer phases. First, the source-like distribution is optimized over several epochs to enhance its quality. This enhanced source-like distribution is then used to transfer dependencies to the target encoder during the transfer phase for effective adaptation. 
For a comprehensive understanding of TemSR's training process, the complete pseudo-code is provided in Algorithm \ref{alg:model_update}.\label{sec:pesudo}
\begin{algorithm}[!b]
\caption{Overall Adaptation Process}
\label{alg:model_update}
\definecolor{codeblack}{rgb}{0.25,0.5,0.5}
\definecolor{codekeyword}{rgb}{0.13,0.13,1} 
\lstset{
  backgroundcolor=\color{white},
  basicstyle=\fontsize{7.2pt}{7.2pt}\ttfamily\selectfont,
  columns=fullflexible,
  breaklines=true,
  captionpos=b,
  commentstyle=\fontsize{7.2pt}{7.2pt}\color{codeblack},
  keywordstyle=\fontsize{7.2pt}{7.2pt}\color{codekeyword}, 
  language=Python,
  morekeywords={def, in},
}

\begin{lstlisting}[language=python]
# X_T, target sample [N, L], N: number of sensors, L: time length

# M: masking function
# H: entropy computation function

# F_S: source domain pretrained encoder
# G: source domain pretrained classifier

# F_T: target domain encoder, initialized by F_S
# R: recovery model

# A_B: anchor bank storing recovered samples
# E_B: entropy bank storing entropy for recovered samples

# num_epochs: number of training epochs

F_S.eval()  # Freeze source encoder
G.eval()    # Freeze source classifier
F_T.train() # Trainable target encoder
R.train()   # Trainable recovery model

# Initialize anchor and entropy banks
A_B.initial()
E_B.initial()

for epo in num_epochs:

    # Step 1: Masking and recovery
    X_hat = M(X_T)        # Mask the target sample
    X_Sl = R(X_hat)       # Recover masked target sample
    
    # Step 2: Update anchor and entropy banks
    E_Sl = H(G(F_S(X_Sl)))  # Compute entropy of recovered sample
    A_B.update(X_Sl.detach())        # Update anchor bank with recovered samples
    E_B.update(E_Sl.detach())        # Update entropy bank

    # Step 3: Compute anchor-based recovery diversity maximization (L_ARDM)
    A = A_B.index(top_k(E_B))    # Select top samples by entropy
    L_ARDM = Anchor_Info_Max(X_Sl, X_T, A)

    # Step 4: Compute local context-aware regularization loss (L_LCA)
    L_LCA = Local_Context_Entropy(X_Sl)

    # Step 5: Compute feature alignment loss (L_Align)
    h_Sl = F_S(X_Sl)        # Extract features of source-like samples
    h_T = F_T(X_T)          # Extract features of target samples
    L_Align = Alignment(h_Sl, h_T)  # Align source-like and target features

    # Step 6: Compute target entropy loss (L_TrgEnt)
    L_TrgEnt = H(G(h_T))    # Compute entropy of target prediction

    # Step 7: Cycle between source-like optimization and adaptation
    if epo in source-like optimization phase:
        loss = combine_losses(L_ARDM, L_LCA, L_TrgEnt)  # Source-like optimization
    else:
        loss = combine_losses(L_Align, L_TrgEnt)         # Adaptation

    # Step 8: Backpropagation and optimization
    loss.backward()
    optimizer.step()

\end{lstlisting}
\end{algorithm}

\subsection{Trivial Solutions with Large Masking ratio}
\label{sec:trivial_solution}
We present the proof demonstrating that high masking ratios lead to trivial solutions in source-like sample recovery process.

\begin{theorem}
With a high masking ratio, the recovery model is prone to collapsing to a constant value for the source-like domain, thus impairing the performance of domain adaptation.
\end{theorem}

\textbf{Proof:}

\paragraph{Given Conditions}
\begin{itemize}
    \item $\mathbi{X}_T^i$ is a TS sample from the target domain;
    \item $M(\mathbi{X}_T^i)$ is the masking operation applied to $\mathbi{X}_T^i$, with a masking ratio $ p_m $ representing the masked proportion;
    \item $\mathcal{R}_\zeta$ is the recovery model, parameterized by $\zeta$, which completes the masked sample to generate a source-like sample $\mathbi{X}_{Sl}^i = \mathcal{R}_\zeta(M(\mathbi{X}_T^i))$;
    \item $\mathcal{F}_{\theta}$ is the fixed pretrained source encoder, aiming to extract the feature $\mathbf{z}$ from the recovered sample $\mathbi{X}_{Sl}^i$;
    \item $p(\mathbf{z})$ is the probability distribution of the feature $\mathbf{z}$.
\end{itemize}

The entropy of the feature distribution is given as Eq. (\ref{eq:ent_example}), and the training objective is minimizing this entropy.
\begin{equation}
\label{eq:ent_example}
    H(p(\mathbf{z})) = -\int p(\mathbf{z}) \log p(\mathbf{z}) \, d\mathbf{z}.
\end{equation}
\paragraph{Feature Collapse in High Masking Ratio}
As the masking ratio $p_m$ increases toward 1, the masked sample $M(\mathbi{X}_T^i)$ contains minimal information for the original target data $\mathbi{X}_T^i$. Consequently, the recovery model $\mathcal{R}_\zeta$ faces increasing difficulty in generating meaningful samples. To achieve the training objectives in Eq. (\ref{eq:ent_example}) for entropy minimization, the model may try to find a degenerate solution where the recovered sample $\mathbi{X}_{Sl}^i = \mathcal{R}_\zeta(M(\mathbi{X}_T^i))$ becomes constant across the masked region, as doing so can easily minimize entropy to 0.

Specifically, for a high masking ratio, $\mathbi{X}_{Sl}^i$ is approximated by a constant value $c$, i.e.
\begin{equation}\mathbi{X}_{Sl}^i \approx c \quad \text{with     } p_m \approx 1.\end{equation}
Passing this constant through the encoder results in constant feature representations:
\begin{equation}\mathbf{z} = \mathcal{F}_{\theta}(\mathbi{X}_{Sl}^i) \approx \mathcal{F}_{\theta}(c) = z_0.\end{equation}
In this case, the distribution of $\mathbf{z}$ collapses to a Dirac delta function centered at $z_0$:
\begin{equation}
\label{eq:delta}
p(\mathbf{z}) = \delta(\mathbf{z} - z_0).
\end{equation}
By substituting Eq. (\ref{eq:delta}) into the entropy (\ref{eq:ent_example}) and using the property $\delta(\mathbf{x}) \log \delta(\mathbf{x}) = 0$ for a delta function $p(\mathbf{z}) = \delta(\mathbf{z} - z_0)$, we derive the entropy of the collapsed features:
\begin{equation}H(p(\mathbf{z})) = -\int \delta(\mathbf{z} - z_0) \log \delta(\mathbf{z} - z_0) \, d\mathbf{z} = 0.\end{equation}

This implies that the entropy $H(p(\mathbf{z}))$ reaches its minimum value of zero, which satisfies the optimization objective but results in feature collapse. The model converges to a trivial solution where no meaningful variability in the recovered source-like sample exists.

\paragraph{Conclusion}
Given the high masking ratio, the recovery model $\mathcal{R}_\zeta$ is unable to generate a valid source-like sample. Instead, it defaults to generating a constant value to minimize the entropy, resulting in collapsed features that carry no useful information. This trivial solution, characterized by $p(\mathbf{z}) = \delta(\mathbf{z} - z_0)$, leads to zero entropy, while the recovered sample fails to capture the temporal dependencies required for successful domain adaptation. In contrast, a lower masking ratio provides the recovery model with sufficient context, allowing for more meaningful generations. When paired with our designed anchor-based recovery diversity maximization module, this results in diverse, temporally coherent recovered samples. Thus, a lower masking ratio, in conjunction with diversity-enhancing techniques, is critical to ensuring effective recovery and adaptation.

\subsection{Improved Diversity with Recovery Diversity Maximization}
\label{sec:diversity}
We present the proof demonstrating that maximizing the distance between the original and recovered samples enhances the diversity of the recovered distribution.

\begin{theorem}
Maximizing the distance between original samples $\mathbi{X}_T^i$ and recovered samples $\mathbi{X}_{Sl}^i$ enhances the diversity of the recovered samples.
\end{theorem}

\textbf{Proof:}

\paragraph{Given Conditions}

\begin{itemize}
    \item $ \mathbi{X}_T^i $ is a TS sample from the target domain.
    \item $ \mathbi{X}_{Sl}^i $ is the corresponding recovered sample, generated by the recovery model $ \mathcal{R}_\zeta $, i.e., $ \mathbi{X}_{Sl}^i = \mathcal{R}_\zeta(M(\mathbi{X}_T^i)) $, where $ M(\mathbi{X}_T^i) $ is the masked version of $ \mathbi{X}_T^i $.
    \item $ p(\mathbi{X}_T^i, \mathbi{X}_{Sl}^i) $ denotes the joint probability distribution of the original samples $ \mathbi{X}_T^i $ and recovered samples $ \mathbi{X}_{Sl}^i $.
    \item $ d(\mathbi{X}_T^i, \mathbi{X}_{Sl}^i) $ is the distance between the original and recovered samples.
\end{itemize}
\paragraph{Conditional Entropy and Diversity}

The conditional entropy $ H(\mathbi{X}_{Sl}^i | \mathbi{X}_T^i) $ measures the uncertainty in the recovered sample $ \mathbi{X}_{Sl}^i $, given the original sample $ \mathbi{X}_T^i $. As $ \mathbi{X}_{Sl}^i = \mathcal{R}_\zeta(M(\mathbi{X}_T^i)) $, higher conditional entropy implies greater uncertainty of $ \mathbi{X}_{Sl}^i $ generated from $ \mathbi{X}_T^i $, suggesting a wider range of possible outcomes for the recovered samples from their original samples. Thus, increasing the conditional entropy directly corresponds to enhancing the recovered samples' diversity.

\paragraph{Conditional Entropy Equation}

The conditional entropy $ H(\mathbi{X}_{Sl}^i | \mathbi{X}_T^i) $ quantifies the uncertainty in $ \mathbi{X}_{Sl}^i $, given $ \mathbi{X}_T^i $, and is defined as:
\begin{equation}
\label{eq:cond_ent}
H(\mathbi{X}_{Sl}^i | \mathbi{X}_T^i) = - \sum_{\mathbi{X}_T^i} \sum_{\mathbi{X}_{Sl}^i} p(\mathbi{X}_T^i, \mathbi{X}_{Sl}^i) \log p(\mathbi{X}_{Sl}^i | \mathbi{X}_T^i).
\end{equation}
This equation measures how much uncertainty remains in $ \mathbi{X}_{Sl}^i $ after observing $ \mathbi{X}_T^i $. Higher values of $ H(\mathbi{X}_{Sl}^i | \mathbi{X}_T^i) $ indicate greater diversity in the recovered samples.

\paragraph{Probability Decay with Distance}

We now show that the joint probability $ p(\mathbi{X}_T^i, \mathbi{X}_{Sl}^i) $ is inversely related to the distance $ d(\mathbi{X}_T^i, \mathbi{X}_{Sl}^i) $. Intuitively, nearby events have higher probabilities, while distant events have lower probabilities. For example, in a Gaussian distribution, the probability density decays as the distance between $ \mathbi{X}_T^i $ and $ \mathbi{X}_{Sl}^i $ increases:
\begin{equation}
\label{eq:example_gaussian}
p(\mathbi{X}_T^i, \mathbi{X}_{Sl}^i) \propto \exp\left( -\frac{d(\mathbi{X}_T^i, \mathbi{X}_{Sl}^i)^2}{2\sigma^2} \right).
\end{equation}
Here, $ d(\mathbi{X}_T^i, \mathbi{X}_{Sl}^i) $ is the distance between the original and recovered samples, and $ \sigma^2 $ is the variance. As $ d(\mathbi{X}_T^i, \mathbi{X}_{Sl}^i) $ increases, the probability $ p(\mathbi{X}_T^i, \mathbi{X}_{Sl}^i) $ decays exponentially.

Based on Eq. (\ref{eq:cond_ent}) and Eq. (\ref{eq:example_gaussian}), we can derive the chain of relationships of (\ref{eq:chain_relation}): as the distance $ d(\mathbi{X}_T^i, \mathbi{X}_{Sl}^i) $ increases, the conditional entropy $H(\mathbi{X}_{Sl}^i | \mathbi{X}_T^i)$ from Eq. (\ref{eq:cond_ent}) also increases, indicating the larger diversity in $\mathbi{X}_{Sl}^i $ given $ \mathbi{X}_T^i $. 
\begin{equation}
\label{eq:chain_relation}
d(\mathbi{X}_T^i, \mathbi{X}_{Sl}^i) \uparrow \quad \Rightarrow p(\mathbi{X}_T^i, \mathbi{X}_{Sl}^i) \downarrow \quad \Rightarrow H(\mathbi{X}_{Sl}^i | \mathbi{X}_T^i) \uparrow.
\end{equation}
\paragraph{Conclusion}

Maximizing $ d(\mathbi{X}_T^i, \mathbi{X}_{Sl}^i) $ decreases the joint probability $ p(\mathbi{X}_T^i, \mathbi{X}_{Sl}^i) $, thus increasing the conditional entropy $ H(\mathbi{X}_{Sl}^i | \mathbi{X}_T^i) $. As higher conditional entropy corresponds to greater diversity in the recovered samples, we conclude that maximizing the distance between the original and recovered samples enhances the diversity of the recovered distribution.

\begin{table*}[htbp!]
  \centering
  \caption{Detailed results of the ten HAR cross-domain scenarios in terms of MF1 score (\%).}
  \begin{adjustbox}{width = 1.0\linewidth,center}
    \begin{tabular}{lcccccccccccc}
    \toprule
    \toprule
    Models & SF    & 2$\rightarrow$11 & 12$\rightarrow$16 & 9$\rightarrow$18 & 6$\rightarrow$23 & 7$\rightarrow$13 & 18$\rightarrow$27 & 20$\rightarrow$5 & 24$\rightarrow$8 & 28$\rightarrow$27 & 30$\rightarrow$20 & AVG \\
    \midrule
    SRC & \bmark & 95.69$\pm$5.72 & 67.13$\pm$9.83 & 70.07$\pm$4.71 & 81.01$\pm$14.9 & 84.5$\pm$12.08 & 85.95$\pm$5.00 & 63.30$\pm$4.13 & 71.59$\pm$8.56 & 50.24$\pm$5.92 & 67.91$\pm$9.21 & 73.73$\pm$2.68 \\
    TRG & \bmark & 100.0$\pm$0.00 & 98.50$\pm$1.30 & 100.0$\pm$0.00 & 100.0$\pm$0.00 & 100.0$\pm$0.00 & 100.0$\pm$0.00 & 97.21$\pm$3.08 & 100.0$\pm$0.00 & 100.0$\pm$0.00 & 88.61$\pm$9.36 & 98.43$\pm$2.84 \\
    \midrule
    DANN  & \xmark & 98.09$\pm$1.68 & 62.08$\pm$1.69 & 70.7$\pm$11.36 & 85.6$\pm$15.71 & 93.33$\pm$0.00 & \textbf{100.0$\pm$0.00} & 78.41$\pm$7.67 & 87.99$\pm$9.41 & 97.47$\pm$1.00 & \textbf{87.25$\pm$0.81} & 86.09$\pm$4.86 \\
    CDAN  & \xmark & 98.19$\pm$1.57 & 61.20$\pm$3.27 & 71.3$\pm$14.64 & 96.73$\pm$0.00 & 93.33$\pm$0.00 & \underline{99.61$\pm$0.67} & 82.02$\pm$5.43 & \underline{98.59$\pm$2.44} & \underline{99.12$\pm$1.52} & 80.70$\pm$7.43 & 88.07$\pm$1.22 \\
    CoDATs & \xmark & 86.65$\pm$4.28 & 61.03$\pm$2.33 & 80.51$\pm$8.47 & 92.08$\pm$4.39 & 92.61$\pm$0.51 & 97.67$\pm$1.02 & 82.81$\pm$7.05 & 94.69$\pm$1.81 & 92.29$\pm$9.25 & 80.44$\pm$5.04 & 86.07$\pm$2.88 \\
    CLUDA & \xmark & 80.33$\pm$3.81 & 66.67$\pm$2.24 & 70.35$\pm$2.13 & 91.14$\pm$1.70 & 95.28$\pm$2.62 & \textbf{100.0$\pm$0.00} & 80.73$\pm$3.24 & 91.67$\pm$3.15 & 98.96$\pm$1.47 & 80.43$\pm$2.34 & 85.55$\pm$1.24 \\
    RAINCOAT & \xmark & \textbf{100.0$\pm$0.00} & \textbf{76.28$\pm$3.18} & 77.35$\pm$3.70 & 98.14$\pm$1.20 & \textbf{100.0$\pm$0.00} & \textbf{100.0$\pm$0.00} & \textbf{85.73$\pm$3.02} & 97.67$\pm$2.31 & \textbf{100.0$\pm$0.00} & \underline{86.46$\pm$1.04} & \textbf{92.16$\pm$0.83} \\
    \midrule
    SHOT  & \cmark & \textbf{100.0$\pm$0.00} & 70.76$\pm$6.22 & 70.19$\pm$8.99 & \underline{98.91$\pm$1.89} & 93.01$\pm$0.57 & 92.93$\pm$2.79 & 69.66$\pm$1.06 & 88.58$\pm$3.94 & 90.39$\pm$3.11 & 75.47$\pm$1.96 & 84.99$\pm$2.00 \\
    NRC   & \cmark & 97.02$\pm$2.82 & \underline{72.18$\pm$0.59} & 63.10$\pm$4.84 & 96.41$\pm$1.33 & 89.13$\pm$0.54 & \textbf{100.0$\pm$0.00} & 81.82$\pm$1.19 & 92.97$\pm$3.21 & 98.43$\pm$0.88 & 82.97$\pm$2.71 & 87.40$\pm$0.34 \\
    AaD   & \cmark & 98.51$\pm$2.58 & 66.15$\pm$6.15 & 68.33$\pm$11.9 & 98.07$\pm$1.71 & 89.41$\pm$2.86 & \textbf{100.0$\pm$0.00} & 80.75$\pm$2.72 & 94.69$\pm$3.42 & 84.85$\pm$13.1 & 77.77$\pm$1.43 & 85.85$\pm$1.29 \\
    BAIT  & \cmark & \underline{98.88$\pm$1.93} & 56.65$\pm$2.54 & 80.4$\pm$13.43 & \textbf{100.0$\pm$0.00} & 97.43$\pm$3.59 & \textbf{100.0$\pm$0.00} & 80.91$\pm$1.60 & \textbf{100.0$\pm$0.00} & \textbf{100.0$\pm$0.00} & 82.66$\pm$1.30 & 89.69$\pm$1.23 \\
    SF(DA)$^2$  & \cmark & \textbf{100.0$\pm$0.00} & 63.94$\pm$2.78 & 72.87$\pm$9.23 & 96.73$\pm$1.79 & 93.33$\pm$1.44 & \textbf{100.0$\pm$0.00} & 81.06$\pm$0.60 & 98.88$\pm$1.92 & \textbf{100.0$\pm$0.00} & 83.63$\pm$8.70 & 89.04$\pm$1.61 \\
    MAPU  & \cmark & \textbf{100.0$\pm$0.00} & 67.96$\pm$4.62 & 82.77$\pm$2.54 & 97.82$\pm$1.89 & \underline{99.29$\pm$1.22} & \textbf{100.0$\pm$0.00} & \underline{82.88$\pm$3.68} & 96.48$\pm$3.09 & 96.01$\pm$3.19 & 85.43$\pm$3.84 & 90.86$\pm$0.98 \\
    TFDA  & \cmark & \textbf{100.0$\pm$0.00} & 65.44$\pm$2.56 & \underline{86.67$\pm$3.21} & 90.17$\pm$2.78 & 93.89$\pm$8.10 & \textbf{100.0$\pm$0.00} & 79.89$\pm$0.78 & \textbf{100.0$\pm$0.00} & \textbf{100.0$\pm$0.00} & 80.26$\pm$4.61 & 89.63$\pm$0.56 \\
    \midrule
    TemSR & \cmark & \textbf{100.0$\pm$0.00} & 64.21$\pm$3.04 & \textbf{93.65$\pm$2.02} & 97.82$\pm$1.89 & 98.95$\pm$0.01 & \textbf{100.0$\pm$0.00} & 82.32$\pm$0.73 & \textbf{100.0$\pm$0.00} & \textbf{100.0$\pm$0.00} & 84.10$\pm$5.52 & \underline{92.10$\pm$0.33} \\
    \bottomrule
    \bottomrule
    \end{tabular}%
    \end{adjustbox}
  \label{tab:har}%
\end{table*}%
\begin{table*}[htbp!]
  \centering
  \caption{Detailed results of the ten SSC cross-domain scenarios in terms of MF1 score (\%).}
  \begin{adjustbox}{width = 1.0\linewidth,center}
    \begin{tabular}{lcccccccccccl}
    \toprule
    \toprule
    Models & SF    & 16$\rightarrow$1 & 9$\rightarrow$14 & 12$\rightarrow$5 & 7$\rightarrow$18 & 0$\rightarrow$11 & 3$\rightarrow$19 & 18$\rightarrow$12 & 13$\rightarrow$17 & 5$\rightarrow$15 & 6$\rightarrow$2 & \multicolumn{1}{c}{AVG} \\
    \midrule
    SRC & \bmark & 52.93$\pm$3.42 & 63.99$\pm$8.04 & 48.79$\pm$3.31 & 62.33$\pm$3.86 & 50.43$\pm$6.26 & 47.38$\pm$3.36 & 38.35$\pm$2.03 & 43.80$\pm$0.12 & 60.13$\pm$6.36 & 55.67$\pm$2.20 & 52.38$\pm$0.47 \\
    TRG & \bmark & 81.52$\pm$2.06 & 75.79$\pm$0.88 & 73.87$\pm$1.43 & 77.74$\pm$1.86 & 68.26$\pm$0.73 & 78.79$\pm$1.49 & 73.51$\pm$1.73 & 70.39$\pm$0.75 & 72.17$\pm$1.99 & 82.11$\pm$1.13 & 75.41$\pm$0.43 \\
    \midrule
    DANN  & \xmark & 58.68$\pm$3.29 & 64.29$\pm$1.08 & 64.65$\pm$1.83 & 69.54$\pm$3.00 & 44.13$\pm$5.84 & 64.09$\pm$4.48 & \underline{54.33$\pm$4.81} & 52.31$\pm$1.70 & 68.03$\pm$0.29 & \underline{71.78$\pm$2.24} & 61.18$\pm$2.31 \\
    CDAN  & \xmark & 59.65$\pm$4.96 & 64.18$\pm$6.37 & 64.43$\pm$1.17 & 67.61$\pm$3.55 & 39.38$\pm$3.28 & 60.19$\pm$1.16 & 40.46$\pm$6.79 & 40.82$\pm$8.87 & 65.22$\pm$6.73 & 68.81$\pm$1.86 & 57.07$\pm$1.79 \\
    CoDATs & \xmark & \underline{63.84$\pm$3.36} & 63.51$\pm$6.92 & 52.54$\pm$5.94 & 66.06$\pm$2.48 & 46.28$\pm$5.99 & 66.15$\pm$4.46 & 47.84$\pm$5.59 & 38.17$\pm$10.8 & \underline{72.62$\pm$3.07} & 61.59$\pm$13.1 & 57.86$\pm$0.76 \\
    CLUDA & \xmark & 55.67$\pm$1.21 & 64.33$\pm$1.24 & 60.12$\pm$4.55 & 64.35$\pm$1.55 & 46.78$\pm$2.55 & 64.33$\pm$2.22 & 45.56$\pm$1.34 & 51.12$\pm$6.77 & 64.55$\pm$1.21 & 61.12$\pm$3.34 & 57.79$\pm$1.37 \\
    RAINCOAT & \xmark & 59.04$\pm$2.02 & 68.04$\pm$1.18 & 62.20$\pm$3.22 & 66.77$\pm$1.56 & 49.17$\pm$2.70 & \textbf{68.89$\pm$0.66} & 49.40$\pm$1.25 & 50.71$\pm$6.68 & \textbf{73.53$\pm$0.51} & \textbf{72.09$\pm$2.38} & \underline{61.98$\pm$1.48} \\
    \midrule
    SHOT  & \cmark & 59.07$\pm$2.14 & 69.93$\pm$0.46 & 62.11$\pm$1.62 & 69.74$\pm$1.22 & \textbf{50.78$\pm$1.90} & 65.44$\pm$1.06 & 48.14$\pm$11.2 & 56.41$\pm$1.60 & 55.51$\pm$9.37 & 64.56$\pm$2.16 & 60.16$\pm$3.82 \\
    NRC   & \cmark & 52.09$\pm$1.89 & 58.52$\pm$0.66 & 59.87$\pm$2.48 & 66.18$\pm$0.25 & 47.55$\pm$1.72 & 64.65$\pm$2.25 & 52.86$\pm$6.60 & 56.93$\pm$2.89 & 61.89$\pm$5.94 & 66.54$\pm$2.29 & 58.70$\pm$2.79 \\
    AaD   & \cmark & 57.04$\pm$2.03 & 65.27$\pm$1.69 & 61.84$\pm$1.74 & 67.35$\pm$1.48 & 44.04$\pm$2.18 & 52.42$\pm$4.55 & 40.86$\pm$8.43 & \textbf{58.28$\pm$6.97} & 63.06$\pm$12.3 & 59.29$\pm$2.90 & 56.94$\pm$3.52 \\
    BAIT  & \cmark & 56.83$\pm$1.17 & 71.84$\pm$1.18 & \underline{65.57$\pm$2.15} & 71.12$\pm$1.45 & 42.30$\pm$2.61 & 59.56$\pm$1.87 & 53.53$\pm$1.89 & 53.03$\pm$3.53 & 60.53$\pm$5.08 & 63.69$\pm$1.04 & 59.80$\pm$0.60 \\
    SF(DA)$^2$  & \cmark & 64.92$\pm$0.38 & 71.95$\pm$0.44 & 46.18$\pm$0.42 & 64.20$\pm$0.06 & 39.13$\pm$1.78 & 51.31$\pm$4.60 & 49.63$\pm$3.11 & 43.85$\pm$1.56 & 62.45$\pm$12.3 & 60.89$\pm$4.44 & 55.45$\pm$0.86 \\
    MAPU  & \cmark & \textbf{63.85$\pm$4.63} & \textbf{74.73$\pm$0.64} & 64.08$\pm$2.21 & \textbf{74.21$\pm$0.58} & 43.36$\pm$5.49 & 59.03$\pm$3.60 & 52.82$\pm$4.94 & 48.09$\pm$2.25 & 67.04$\pm$1.22 & 58.98$\pm$1.07 & 60.61$\pm$1.28 \\
    TFDA  & \cmark & 63.67$\pm$1.62 & 71.64$\pm$1.46 & 62.87$\pm$3.78 & 69.99$\pm$2.01 & 43.89$\pm$8.34 & 56.23$\pm$1.33 & 51.48$\pm$1.67 & 55.77$\pm$3.65 & 59.78$\pm$1.98 & 59.89$\pm$2.48 & 59.52$\pm$0.78 \\

    \midrule   
    
    \textbf{TemSR} & \cmark & 62.51$\pm$1.09 & \underline{72.60$\pm$0.74} & \textbf{66.70$\pm$1.91} & \underline{72.15$\pm$1.01} & \underline{49.62$\pm$1.88} & \underline{65.87$\pm$0.53} & \textbf{60.32$\pm$0.97} & \underline{57.56$\pm$2.07} & 66.50$\pm$2.07 & 64.82$\pm$1.78 & \textbf{63.86$\pm$0.58} \\
    \bottomrule
    \bottomrule
    \end{tabular}%
    \end{adjustbox}
  \label{tab:ssc}%
\end{table*}%

\begin{table*}[htbp!]
  \centering
  \caption{Detailed results of the ten MFD cross-domain scenarios in terms of MF1 score (\%).}
  \begin{adjustbox}{width = 1.0\linewidth,center}
    \begin{tabular}{lcccccccccccl}
    \toprule
    \toprule
    Models & SF    & 0$\rightarrow$1 & 1$\rightarrow$0 & 1$\rightarrow$2 & 2$\rightarrow$3 & 3$\rightarrow$1 & 0$\rightarrow$3 & 1$\rightarrow$3 & 2$\rightarrow$1 & 3$\rightarrow$0 & 3$\rightarrow$2 & \multicolumn{1}{c}{AVG} \\
    \midrule
    SRC & \bmark & 26.26$\pm$5.04 & 68.63$\pm$6.22 & 72.66$\pm$0.95 & 96.90$\pm$1.38 & 99.02$\pm$1.07 & 42.13$\pm$8.06 & 96.25$\pm$3.72 & 86.96$\pm$0.58 & 46.42$\pm$2.42 & 71.71$\pm$6.54 & 70.69$\pm$2.61 \\
    TRG & \bmark & 100.0$\pm$0.00 & 97.88$\pm$1.60 & 99.92$\pm$0.14 & 100.0$\pm$0.00 & 100.0$\pm$0.00 & 100.0$\pm$0.00 & 100.0$\pm$0.00 & 100.0$\pm$0.00 & 97.88$\pm$1.60 & 99.92$\pm$0.14 & 99.56$\pm$2.31 \\
    \midrule
    DANN  & \xmark & 83.44$\pm$1.72 & 51.52$\pm$0.38 & 84.19$\pm$2.10 & \textbf{99.95$\pm$0.09} & \textbf{100.0$\pm$0.00} & 77.65$\pm$9.41 & \underline{99.97$\pm$0.04} & \textbf{99.75$\pm$0.14} & 50.85$\pm$1.74 & 72.32$\pm$22.3 & 81.96$\pm$2.89 \\
    CDAN  & \xmark & 84.97$\pm$0.62 & 52.39$\pm$0.49 & \underline{85.96$\pm$0.90} & \underline{99.70$\pm$0.45} & \textbf{100.0$\pm$0.00} & 85.38$\pm$0.42 & \textbf{100.0$\pm$0.00} & \underline{99.02$\pm$0.90} & 62.17$\pm$6.32 & 79.76$\pm$2.75 & 84.93$\pm$1.47 \\
    CoDATs & \xmark & 67.42$\pm$13.3 & 49.92$\pm$13.7 & \textbf{89.05$\pm$4.73} & 99.21$\pm$0.79 & 99.92$\pm$0.14 & 55.68$\pm$3.07 & 99.95$\pm$0.09 & \textbf{99.75$\pm$0.29} & 51.77$\pm$1.86 & \underline{83.36$\pm$1.25} & 79.60$\pm$1.27 \\
    CLUDA & \xmark & 84.43$\pm$1.43 & 55.66$\pm$5.76 & 81.12$\pm$1.20 & 91.13$\pm$1.32 & 93.44$\pm$1.26 & 89.94$\pm$2.33 & 97.12$\pm$0.98 & 91.23$\pm$0.88 & 73.35$\pm$3.44 & 79.98$\pm$6.67 & 83.74$\pm$1.32 \\
    RAINCOAT & \xmark & 88.09$\pm$1.40 & 59.41$\pm$6.61 & 83.87$\pm$0.69 & 93.67$\pm$1.15 & 94.95$\pm$0.71 & 91.19$\pm$0.95 & 97.73$\pm$0.84 & 92.53$\pm$0.79 & 78.45$\pm$2.84 & \textbf{84.61$\pm$0.95} & 86.45$\pm$1.12 \\
    \midrule
    SHOT  & \cmark & 41.99$\pm$2.78 & 57.00$\pm$0.09 & 80.70$\pm$1.49 & 99.48$\pm$0.31 & 99.95$\pm$0.05 & 83.63$\pm$2.32 & 89.33$\pm$3.50 & 88.98$\pm$1.59 & 72.89$\pm$7.84 & 71.38$\pm$2.31 & 78.53$\pm$1.98 \\
    NRC   & \cmark & 73.99$\pm$1.36 & 74.88$\pm$8.81 & 69.23$\pm$0.75 & 78.04$\pm$11.3 & 71.48$\pm$4.59 & 70.88$\pm$1.75 & 70.35$\pm$6.80 & 72.10$\pm$1.34 & 63.67$\pm$5.57 & 61.52$\pm$3.20 & 70.61$\pm$1.60 \\
    AaD   & \cmark & 71.72$\pm$3.96 & 75.33$\pm$4.65 & 78.31$\pm$2.26 & 90.07$\pm$7.02 & 87.45$\pm$11.7 & 89.35$\pm$2.22 & \textbf{100.0$\pm$0.00} & 96.49$\pm$3.04 & 72.42$\pm$4.47 & 74.56$\pm$6.80 & 83.57$\pm$2.46 \\
    BAIT  & \cmark & 83.1$\pm$14.69 & 60.51$\pm$6.43 & 75.9$\pm$12.51 & 95.57$\pm$2.85 & \textbf{100.0$\pm$0.00} & 82.12$\pm$15.5 & \textbf{100.0$\pm$0.00} & 85.12$\pm$1.49 & 67.21$\pm$3.33 & 83.37$\pm$6.34 & 83.29$\pm$4.60 \\
    
    SF(DA)$^2$  & \cmark & \textbf{100.0$\pm$0.00} & 77.46$\pm$1.23 & 85.26$\pm$1.03 & 94.97$\pm$0.04 & 98.50$\pm$2.59 & \textbf{99.97$\pm$0.04} & \textbf{100.0$\pm$0.00} & 91.17$\pm$1.52 & 80.57$\pm$4.98 & 81.84$\pm$1.91 & 90.97$\pm$0.90 \\
    MAPU  & \cmark & 99.43$\pm$0.51 & 77.42$\pm$0.16 & 85.78$\pm$7.38 & 99.67$\pm$0.50 & \underline{99.97$\pm$0.05} & 85.63$\pm$2.44 & \textbf{100.0$\pm$0.00} & 94.38$\pm$0.62 & \textbf{88.47$\pm$1.99} & 81.51$\pm$2.43 & \underline{91.22$\pm$1.08} \\
    TFDA  & \cmark & 96.78$\pm$2.13 & \underline{83.66$\pm$7.24} & 82.79$\pm$5.41 & 97.37$\pm$2.11 & \textbf{100.0$\pm$0.00} & 97.15$\pm$0.12 & \textbf{100.0$\pm$0.00} & 90.48$\pm$5.01 & 82.77$\pm$1.77 & 75.91$\pm$4.89 & 90.69$\pm$2.48 \\
    \midrule
    \textbf{TemSR} & \cmark & \underline{99.97$\pm$0.02} & \textbf{87.03$\pm$4.05} & 84.47$\pm$5.88 & 95.23$\pm$3.85 & \textbf{100.0$\pm$0.00} & \underline{99.95$\pm$0.05} & \textbf{100.0$\pm$0.00} & 96.67$\pm$4.21 & \underline{87.17$\pm$1.56} & 81.96$\pm$5.09 & \textbf{93.24$\pm$1.83} \\
    \bottomrule
    \bottomrule
    \end{tabular}%
    \end{adjustbox}
  \label{tab:fd}%
\end{table*}%

\begin{table*}[htbp!]
  \centering
  \caption{Detailed results of the ten HHAR cross-domain scenarios in terms of MF1 score (\%).}
  \begin{adjustbox}{width = 1.0\linewidth,center}
    \begin{tabular}{lcccccccccccc}
    \toprule
    \toprule
    Models & SF    & 0$\rightarrow$6   & 1$\rightarrow$6   & 2$\rightarrow$7   & 3$\rightarrow$8   & 4$\rightarrow$5   & 5$\rightarrow$0   & 6$\rightarrow$1   & 7$\rightarrow$4   & 8$\rightarrow$3   & 0$\rightarrow$2   & \multicolumn{1}{c}{AVG} \\
    \midrule
    SRC   & \bmark & 54.50$\pm$4.13 & 80.05$\pm$7.31 & 39.13$\pm$4.11 & 75.60$\pm$4.69 & 83.79$\pm$7.33 & 22.80$\pm$4.07 & 75.09$\pm$10.7 & 74.83$\pm$10.22 & 69.25$\pm$7.40 & 46.14$\pm$10.1 & 62.11$\pm$2.50 \\
    TRG   & \bmark & 95.85$\pm$1.87 & 95.85$\pm$1.87 & 96.89$\pm$4.34 & 98.60$\pm$0.15 & 97.86$\pm$0.68 & 96.91$\pm$1.27 & 98.35$\pm$0.48 & 97.07$\pm$0.87 & 98.50$\pm$0.27 & 90.76$\pm$11.1 & 96.66$\pm$0.63 \\
    \midrule
    DANN  & \xmark & 45.59$\pm$0.37 & 92.21$\pm$0.38 & 44.43$\pm$0.61 & 90.61$\pm$8.77 & 97.06$\pm$1.43 & 29.92$\pm$4.98 & \textbf{93.95$\pm$1.25} & 94.62$\pm$1.81 & 89.37$\pm$1.36 & \textbf{77.08$\pm$3.16} & 75.48$\pm$1.85 \\
    CDAN  & \xmark & 45.56$\pm$1.56 & 93.40$\pm$1.08 & 49.46$\pm$8.46 & 86.77$\pm$9.54 & \textbf{97.43$\pm$0.74} & \textbf{39.88$\pm$6.59} & 91.97$\pm$2.78 & 93.87$\pm$3.10 & \textbf{97.24$\pm$0.27} & 74.65$\pm$10.1 & 77.02$\pm$2.65 \\
    CoDATS & \xmark & 50.86$\pm$5.81 & 93.03$\pm$0.42 & 47.47$\pm$9.71 & 94.56$\pm$3.95 & 91.79$\pm$4.81 & \underline{39.42$\pm$6.12} & \underline{93.23$\pm$2.04} & 92.25$\pm$1.20 & 96.65$\pm$0.78 & 70.92$\pm$11.4 & 77.02$\pm$0.97 \\
    CLUDA & \xmark & 43.75$\pm$6.41 & 86.65$\pm$3.44 & 42.44$\pm$1.94 & 77.69$\pm$3.22 & 89.07$\pm$1.66 & 17.89$\pm$2.81 & 87.21$\pm$1.78 & 85.05$\pm$1.41 & 89.23$\pm$0.94 & 59.30$\pm$2.43 & 67.83$\pm$0.67 \\
    RAINCOAT & \xmark & 44.67$\pm$5.92 & 87.23$\pm$3.16 & 43.23$\pm$1.80 & 82.03$\pm$3.16 & 95.22$\pm$1.10 & 22.43$\pm$3.35 & 87.25$\pm$1.80 & 86.66$\pm$1.28 & 96.26$\pm$0.74 & 67.87$\pm$1.99 & 71.28$\pm$0.45 \\
    \midrule
    SHOT  & \cmark & 58.61$\pm$4.22 & 77.77$\pm$7.51 & 41.53$\pm$6.85 & 89.79$\pm$10.2 & 71.42$\pm$4.97 & 21.31$\pm$6.10 & 69.39$\pm$9.57 & 78.48$\pm$1.52 & 81.59$\pm$1.09 & 68.34$\pm$3.53 & 65.82$\pm$3.71 \\
    NRC   & \cmark & 50.35$\pm$3.73 & 55.48$\pm$5.17 & 43.53$\pm$8.18 & 53.01$\pm$9.10 & 81.82$\pm$6.49 & 37.35$\pm$8.27 & 66.03$\pm$1.68 & 79.22$\pm$5.50 & 74.61$\pm$1.82 & 69.62$\pm$1.81 & 61.10$\pm$3.33 \\
    AaD   & \cmark & 61.99$\pm$2.65 & 91.64$\pm$3.28 & 53.66$\pm$5.14 & 93.27$\pm$5.37 & 96.48$\pm$0.53 & 22.09$\pm$5.49 & 87.02$\pm$0.31 & 94.99$\pm$1.94 & 96.68$\pm$0.25 & 74.74$\pm$8.99 & 77.25$\pm$0.75 \\
    BAIT  & \cmark & 47.92$\pm$5.49 & 79.83$\pm$2.58 & 46.26$\pm$8.94 & 86.33$\pm$3.37 & 87.36$\pm$1.27 & 20.51$\pm$5.76 & 62.99$\pm$0.92 & 80.98$\pm$1.56 & 85.88$\pm$2.11 & 67.39$\pm$12.9 & 66.54$\pm$4.45 \\
    SF(DA)$^2$ & \cmark & 61.96$\pm$1.84 & \underline{93.41$\pm$1.05} & 57.06$\pm$6.48 & 78.38$\pm$2.16 & 94.38$\pm$0.92 & 33.59$\pm$2.98 & 90.60$\pm$3.31 & 92.91$\pm$0.13 & 96.29$\pm$1.01 & 73.75$\pm$6.35 & 77.23$\pm$1.91 \\
    MAPU  & \cmark & \underline{64.41$\pm$6.10} & 92.29$\pm$1.90 & 63.37$\pm$7.69 & 97.55$\pm$6.26 & 97.98$\pm$0.32 & 28.04$\pm$9.45 & 83.30$\pm$8.23 & \underline{95.41$\pm$0.44} & 95.15$\pm$0.11 & 70.09$\pm$10.0 & \underline{78.76$\pm$1.97} \\
    TFDA  & \cmark & 59.68$\pm$1.22 & 91.49$\pm$2.32 & \underline{57.88$\pm$9.89} & \underline{92.77$\pm$2.41} & 93.46$\pm$0.92 & 29.94$\pm$5.78 & 79.60$\pm$6.32 & 93.42$\pm$0.88 & 92.34$\pm$1.21 & 69.78$\pm$7.88 & 76.04$\pm$2.17 \\
    \midrule
    TemSR & \cmark & \textbf{64.52$\pm$0.51} & \textbf{94.25$\pm$1.33} & \textbf{59.37$\pm$9.25} & \textbf{98.50$\pm$0.51} & \underline{97.18$\pm$0.18} & 34.84$\pm$4.37 & 78.93$\pm$5.23 & \underline{96.38$\pm$0.20} & \underline{96.92$\pm$0.12} & \underline{76.72$\pm$6.00} & \textbf{79.76$\pm$1.01} \\
    \bottomrule
    \bottomrule
    \end{tabular}%
\end{adjustbox}
  \label{tab:hhar}%
\end{table*}%
\begin{table*}[htbp!]
  \centering
  \caption{Detailed results of the ten WISDM cross-domain scenarios in terms of MF1 score (\%).}
  \begin{adjustbox}{width = 1.0\linewidth,center}
    \begin{tabular}{lcccccccccccc}
    \toprule
    \toprule
    Models & SF    & 7$\rightarrow$18  & 20$\rightarrow$30 & 35$\rightarrow$31 & 17$\rightarrow$23 & 6$\rightarrow$19  & 2$\rightarrow$11  & 33$\rightarrow$12 & 5$\rightarrow$26  & 28$\rightarrow$4  & 23$\rightarrow$32 & \multicolumn{1}{c}{AVG} \\
    \midrule
    SRC   & \bmark & 32.40$\pm$6.88 & 51.81$\pm$2.72 & 20.06$\pm$3.97 & 18.48$\pm$8.30 & 45.82$\pm$15.9 & 43.69$\pm$12.2 & 29.92$\pm$12.5 & 28.41$\pm$1.43 & 52.87$\pm$7.08 & 54.58$\pm$1.70 & 37.80$\pm$2.60 \\
    TRG   & \bmark & 89.32$\pm$7.45 & 96.59$\pm$2.73 & 93.87$\pm$4.78 & 91.72$\pm$1.72 & 98.33$\pm$0.00 & 96.96$\pm$2.70 & 99.28$\pm$0.80 & 90.43$\pm$4.30 & 99.53$\pm$0.80 & 94.46$\pm$4.00 & 95.05$\pm$0.21 \\
    \midrule
    DANN  & \xmark & \underline{53.81$\pm$3.50} & \underline{66.56$\pm$11.6} & 65.06$\pm$6.00 & \textbf{53.74$\pm$12.8} & 64.63$\pm$13.9 & \underline{69.22$\pm$4.86} & 60.11$\pm$3.66 & \textbf{39.81$\pm$12.1} & 62.45$\pm$11.5 & 46.59$\pm$3.83 & \textbf{58.20$\pm$4.39} \\
    CDAN  & \xmark & 46.18$\pm$10.4 & 58.05$\pm$2.39 & 63.05$\pm$13.3 & 34.53$\pm$12.1 & 44.80$\pm$4.03 & 67.82$\pm$13.4 & \underline{60.55$\pm$0.68} & 31.86$\pm$6.32 & \underline{86.24$\pm$14.0} & 44.16$\pm$12.1 & 53.73$\pm$2.56 \\
    CoDATS & \xmark & 39.57$\pm$6.63 & 56.71$\pm$9.14 & 43.44$\pm$2.79 & 50.28$\pm$9.54 & 60.80$\pm$8.03 & 69.51$\pm$7.77 & 60.20$\pm$7.37 & 33.69$\pm$3.38 & \textbf{88.63$\pm$10.9} & 55.54$\pm$5.82 & 55.84$\pm$1.24 \\
    CLUDA & \xmark & 44.62$\pm$4.31 & 55.55$\pm$2.11 & 52.65$\pm$0.66 & 34.62$\pm$3.23 & 56.60$\pm$4.31 & 58.32$\pm$1.01 & 59.37$\pm$3.45 & 27.54$\pm$1.22 & 66.01$\pm$9.33 & 47.65$\pm$5.45 & 50.29$\pm$2.33 \\
    RAINCOAT & \xmark & \textbf{55.96$\pm$3.61} & \textbf{76.47$\pm$1.90} & 52.93$\pm$0.96 & 25.83$\pm$2.29 & 56.62$\pm$9.00 & 68.14$\pm$1.23 & 60.07$\pm$2.32 & 28.26$\pm$2.47 & 66.44$\pm$8.52 & \underline{68.49$\pm$7.01} & 55.92$\pm$2.01 \\
    \midrule
    SHOT  & \cmark & 35.07$\pm$3.16 & 44.19$\pm$5.55 & 47.64$\pm$8.42 & 32.91$\pm$7.03 & 57.24$\pm$5.27 & 51.29$\pm$2.76 & 51.59$\pm$3.70 & 28.71$\pm$1.39 & 51.18$\pm$10.6 & 64.38$\pm$7.25 & 46.42$\pm$2.15 \\
    NRC   & \cmark & 30.60$\pm$3.61 & 54.40$\pm$10.2 & 48.66$\pm$5.87 & 32.39$\pm$7.72 & 61.41$\pm$19.7 & 52.66$\pm$12.3 & 50.18$\pm$2.24 & 29.93$\pm$9.93 & 54.91$\pm$2.97 & 51.40$\pm$17.7 & 46.65$\pm$2.23 \\
    AaD   & \cmark & 48.62$\pm$6.96 & 57.55$\pm$0.91 & 42.26$\pm$11.4 & 42.43$\pm$5.79 & 50.67$\pm$9.34 & 68.60$\pm$0.29 & 43.79$\pm$1.54 & 35.42$\pm$4.71 & 60.12$\pm$1.88 & 67.11$\pm$17.6 & 51.65$\pm$3.51 \\
    BAIT  & \cmark & 30.86$\pm$1.61 & 64.90$\pm$1.72 & \underline{65.24$\pm$6.53} & \underline{53.19$\pm$15.0} & 53.92$\pm$10.5 & 62.87$\pm$2.63 & 42.74$\pm$2.53 & \underline{36.49$\pm$1.96} & 64.27$\pm$3.99 & 65.69$\pm$16.9 & 54.01$\pm$2.85 \\
    SF(DA)$^2$ & \cmark & 45.88$\pm$8.80 & 64.96$\pm$2.32 & 57.52$\pm$0.38 & 36.89$\pm$3.11 & 65.79$\pm$8.94 & 62.56$\pm$9.08 & 43.94$\pm$2.08 & 34.55$\pm$9.11 & 61.26$\pm$0.57 & 67.75$\pm$17.1 & 54.11$\pm$2.44 \\
    MAPU  & \cmark & 39.01$\pm$3.30 & 64.67$\pm$5.42 & 61.25$\pm$8.29 & 39.41$\pm$11.6 & 74.63$\pm$2.06 & 71.09$\pm$2.53 & 58.41$\pm$9.87 & 31.21$\pm$4.48 & 71.98$\pm$5.95 & 65.04$\pm$11.5 & \underline{57.67$\pm$2.82} \\
    TFDA  & \cmark & 37.68$\pm$6.78 & 62.45$\pm$10.1 & 62.33$\pm$7.81 & 39.46$\pm$12.1 & 73.24$\pm$9.77 & 67.44$\pm$10.2 & 57.99$\pm$9.97 & 34.21$\pm$6.76 & 62.34$\pm$4.55 & 67.48$\pm$14.3 & 56.46$\pm$4.53 \\
    \midrule
    TemSR & \cmark & 37.69$\pm$7.28 & 65.49$\pm$9.74 & \textbf{65.57$\pm$6.17} & 43.64$\pm$10.5 & \textbf{78.46$\pm$9.50} & 64.68$\pm$8.98 & \textbf{60.84$\pm$9.98} & 32.28$\pm$2.32 & 64.21$\pm$1.99 & \textbf{69.10$\pm$12.1} & \textbf{58.20$\pm$3.09} \\
    \bottomrule
    \bottomrule
    \end{tabular}%
\end{adjustbox}
  \label{tab:wisdm}%
\end{table*}%
\section{Experiments}
\subsection{Datasets} 
\label{sec:Datasets}
To comprehensively evaluate TemSR, we selected five public TS datasets, including UCI-HAR \cite{uciHAR_dataset}, Sleep-EDF \cite{sleepEDF_dataset}, Machine Fault Diagnosis (MFD) \cite{MFD_dataset}, HHAR \cite{stisen2015smart}, and WISDM \cite{kwapisz2011activity}. The detailed descriptions and preprocessing of these datasets are present as follows.
\subsubsection{UCI-HAR} 
The UCI-HAR dataset is tailored for human activity recognition tasks, comprising sensor data collected from 30 distinct users, each representing a separate domain. Each participant performs six activities: walking, walking upstairs, walking downstairs, standing, sitting, and lying down. The data is recorded using three types of sensors—accelerometers, gyroscopes, and body sensors—each capturing data on three axes. Thus, there are totally nine channels per sample, with each channel containing 128 data points. Following prior research \cite{ragab2023adatime}, we employed a window size of 128 for sample extraction and applied min-max normalization for data preprocessing.

\subsubsection{Sleep-EDF} 
The Sleep-EDF dataset is designed for sleep stage classification. It includes recordings from six channels monitoring various physiological signals, such as EEG (Epz-Cz, Pz-Oz), EOG, and EMG. Based on prior research \cite{MAPU2023ragab} and due to the high information content in the Epz-Cz channel, we utilized only this channel in our experiments. The dataset comprises recordings from 20 subjects, each is treated as a domain because different persons have various personal habits. Each subject can be classified into five sleep stages: wake, light sleep stage 1 (N1), light sleep stage 2 (N2), deep sleep stage 3 (N3), and rapid eye movement (REM) \cite{sleepEDF_dataset}. Notably, each sample in the dataset corresponds to a 30-second window of physiological data, recorded at a sampling rate of 100 Hz, resulting in 3000 timestamps per sample.

\subsubsection{MFD} 
The MFD dataset, collected by Paderborn University, is used for machine fault diagnosis, where vibration signals are collected to identify different types of incipient faults. Data was collected under four distinct working conditions, each treated as a separate domain. Each sample consists of a single univariate channel containing 5120 data points. 

\subsubsection{HHAR}
The HHAR dataset captures sensor measurements obtained from nine participants using both smartphone and smartwatch devices. To ensure consistency across subjects, researchers standardized the data collection process by employing identical Samsung Galaxy S3 smartphones for all participants. The dataset incorporates triaxial sensor measurements from accelerometers, generating three-dimensional TS data per sample with a fixed length of 128 time points per channel. In our experiments, we consider each participant's recordings as an independent domain. Following established protocols from prior work \cite{ragab2023adatime}, we constructed ten distinct cross-domain evaluation scenarios for comprehensive comparisons.

\subsubsection{WISDM}
The WISDM dataset utilizes triaxial accelerometer readings collected from 36 participants, who performed the same six activities as in UCI-HAR. Each sample consists of three-channel TS data, with 128 measurements per axis. Unlike UCI-HAR, this dataset introduces additional complexity due to uneven class distributions across different subjects, posing challenges for model generalization. For domain adaptation, we define each participant’s recordings as an independent domain and construct ten cross-domain scenarios through randomized subject selection, as outlined in prior work \cite{ragab2023adatime}.

\subsection{Unified Training Scheme} 
\label{sec:Unified_Training_Scheme}
To ensure fair comparisons, we utilized a consistent three-layer Convolutional Neural Network (CNN) backbone and adhered to identical training configurations as \cite{MAPU2023ragab}, which is a 1-dimensional CNN comprising three layers with filter sizes of 64, 128, and 128, respectively. Each convolutional layer is followed by a Rectified Linear Unit activation function and batch normalization. Considering potential data imbalances, we used Macro F1-score (MF1) as the primary metric to provide comprehensive evaluations. The mean and standard deviation are reported across three runs for each cross-domain scenario. Full details are provided in Appendix.

\subsection{Comparisons with State-of-the-Arts}
\label{sec:comparison_SOTAs}

For comparisons, we evaluated both conventional UDA methods and SFUDA techniques by following \cite{MAPU2023ragab,nrc,aad}. Conventional UDA methods include DANN \cite{DANN}, CDAN \cite{CDAN}, CoDATS \cite{wilson2020multi}, CLUDA \cite{ozyurt2022contrastive}, and RAINCOAT \cite{raincoat}, while SFUDA methods include SHOT \cite{liang2020we}, NRC \cite{nrc}, AaD \cite{aad}, BAIT \cite{yang2023casting}, SF(DA)$^2$ \cite{hwang2024sfda}, MAPU \cite{MAPU2023ragab}, and TFDA \cite{furqon2025time}. The baselines are introduced in Appendix. Additionally, we report results for source (SRC)-only and target (TRG)-only models to provide the lower and upper bounds of adaptation. For clarity, lower/upper bounds are denoted by \bmark, conventional UDA methods by \xmark, and SFUDA methods by \cmark. Notably, all baselines adopted the same backbone as ours for fairness. Among the SFUDA methods, only MAPU is designed to transfer temporal dependencies across domains, though it requires additional pretraining designs in the source domain.

The performance comparisons on HAR, SSC, MFD, HHAR, and WISDM in Tables \ref{tab:har}, \ref{tab:ssc}, \ref{tab:fd}, \ref{tab:hhar}, and \ref{tab:wisdm}, demonstrate that our method achieves the best performance in most cases. Although it does not outperform all baselines in certain scenarios (e.g., HAR), it is important to note that the top-performing method, RAINCOAT, is a traditional UDA approach that requires access to the source domain during adaptation. In contrast, our method operates in a source-free setting while maintaining competitive performance, underscoring its effectiveness. Among SFUDA methods, the approaches that transfer temporal dependencies across domains—such as MAPU and ours—consistently outperform other SFUDA methods in cross-domain evaluations. On average, MAPU and TemSR achieve the second-best and best results, respectively, across most datasets (e.g., HAR, MFD, HHAR, and WISDM), highlighting the critical role of temporal dependency transfer in TS-SFUDA. Specifically, compared to the best methods that do not consider temporal dependencies (i.e., BAIT, SHOT, SF(DA)$^2$, AaD, and TFDA on five respective datasets), our approach yields significant improvements of 2.41\%, 3.70\%, 2.53\%, 2.51\%, and 1.74\% on these datasets. Even compared with MAPU, our method still improves by 1.24\%, 3.25\%, 2.02\%, 1.00\%, and 0.53\%, respectively. Notably, MAPU relies on source pretraining designs to store temporal dependencies, limiting its practicality. In contrast, our approach adapts entirely in the target domain without any source pretraining designs. Moreover, TemSR effectively recovers the source distribution during adaptation, facilitating a more effective transfer of temporal dependencies and thereby achieving improved and robust performance. These results underscore that without relying on source pretraining designs, TemSR can still transfer temporal dependencies to achieve SOTA performance in TS-SFUDA, even surpassing the existing method that depends on such designs.

\subsection{Ablation Study}
\label{sec:Ablation_Study}
To validate the effectiveness of key components in TemSR, we conduct the ablation study across four analytical dimensions. In the first dimension, we evaluate the overall effects of the recovery and enhancement modules. In the `Baseline (w/o Src-like)' variant, both modules are excluded, and no source-like domain is generated for dependency information transfer. Instead, only the pretrained model is adapted to the target domain via target entropy minimization as \cite{liang2020we}. In the `w/ Recov.' variant, we introduce the recovery module to generate a source-like domain with recovered temporal dependencies for domain adaptation. The `w/ Recov. \& Enhan.' variant further incorporates the enhancement module to improve the diversity of the generated source-like domain, facilitating better recovery. Notably, as the enhancement module builds upon the recovered distribution, it cannot be evaluated independently without the recovery module.

The second dimension examines the components within the recovery module. In the `w/o Local Recov.' variant, local context-aware regularization is replaced with sample-level entropy minimization to regularize the recovery model, aiming to evaluate the importance of recovering local temporal dependencies for effective temporal recovery. In the `w/o Local Consist.' variant, the entropy consistency constraint in Eq. (\ref{eq:siment}) is removed to assess the necessity of ensuring smooth dependencies across different contexts. The third dimension examines the role of local contexts in local temporal recovery. We evaluate three variants: `w/o Early', `w/o Late', and `w/o Recover', each removing a specific local context from the regularization loss to evaluate its individual contribution to recovering local temporal structures. In addition, the `w/o Complete' variant retains only the local contexts while excluding the global context, aiming to assess whether recovering only local temporal structures is sufficient for effective recovery.

The final dimension analyzes the components within the enhancement module. As anchor guidance operates on top of recovery diversity maximization, it cannot be evaluated independently. Moreover, removing the diversity component would revert TemSR to the `w/ Recov. \& Enhan.' variant. Therefore, this dimension specifically focuses on evaluating the anchor guidance mechanism, including three variants: The `w/o Anchor Guid.' variant removes the entire anchor guidance loss, preserving only the recovery diversity maximization. The `w/o Anchor Guid. Aux.' variant retains the main anchor guidance loss but omits the auxiliary guidance loss. Finally, the `w/o Anchor Bank' variant removes the anchor bank and instead generates anchors within each batch, testing whether the anchor bank is essential for producing high-quality anchors. 


\begin{table}[htbp!]
  \centering
  \caption{Ablation study for HAR, SSC, and MFD (\%).}
    \begin{adjustbox}{width = .97\linewidth,center}
    \begin{tabular}{lccc}
    \toprule
    \toprule
    Variants & HAR & SSC & MFD \\
    \midrule
    \multicolumn{4}{c}{Effects of Recovery and Enhancement} \\
    \midrule
    Baseline (w/o Src-like) & 84.99$\pm$2.00 & 60.16$\pm$3.82 & 78.53$\pm$1.98 \\
    w/ Recov. & 90.00$\pm$2.74 & 62.84$\pm$1.44 & 92.46$\pm$2.41 \\
    w/ Recov. \& Enhan. & 92.10$\pm$0.33 & 63.86$\pm$0.58 & 93.24$\pm$1.83 \\
    \midrule
    \midrule
    \multicolumn{4}{c}{Effects in Recovery} \\
    \midrule
    w/o Local Recov. & 90.72$\pm$1.27 & 62.74$\pm$0.81 & 92.33$\pm$2.15 \\
    w/o Local Consist. & 91.50$\pm$0.95 & 63.49$\pm$0.56 & 92.94$\pm$2.67 \\
    \midrule
    \midrule
    \multicolumn{4}{c}{Effects in Local Recovery} \\
    \midrule
    w/o Early & 91.20$\pm$0.81 & 62.97$\pm$1.30 & 92.17$\pm$2.47 \\
    w/o Late & 91.16$\pm$1.35 & 62.95$\pm$1.27 & 92.95$\pm$3.19 \\
    w/o Recover & 91.93$\pm$0.95 & 63.46$\pm$0.37 & 92.96$\pm$0.15 \\
    w/o Complete & 91.04$\pm$0.72 & 62.77$\pm$1.32 & 92.27$\pm$3.35 \\
    \midrule
    \midrule
    \multicolumn{4}{c}{Effects in Enhancement} \\
    \midrule
    w/o Anchor Guid. & 88.91$\pm$1.76 & 62.59$\pm$0.97 & 91.49$\pm$2.34 \\
    w/o Anchor Guid. Aux. & 90.13$\pm$1.68 & 63.43$\pm$0.50 & 92.79$\pm$3.24 \\
    w/o Anchor Bank & 91.97$\pm$0.97 & 63.23$\pm$0.11 & 93.09$\pm$2.32 \\
    \bottomrule
    \bottomrule
    \end{tabular}%
    \end{adjustbox}
  \label{tab:abla}%
\end{table}%
The results in Table \ref{tab:abla} summarize the average performance across all cross-domain cases, with detailed results provided in Appendix. Here, four key insights emerge. First, the source-like domain generated by the recovery and enhancement modules significantly improves performance for TS-SFUDA. In the `Baseline (w/o Src-like)' variant, where both modules are excluded, temporal dependencies cannot be effectively transferred, resulting in the lowest performance. Introducing the recovery module in the `w/ Recov.' variant leads to substantial improvement—e.g., a 5.01\% increase on HAR—demonstrating the effectiveness of generating a source-like domain for transferring temporal dependencies. Furthermore, the `w/ Recov. \& Enhan.' variant, which additionally incorporates the enhancement module, achieves further gains—e.g., an additional 2.01\% on HAR—resulting in the best overall performance. This highlights the critical role of enhancing diversity in the recovered distribution for a more effective source-like domain.

Second, we observe the effectiveness of the components within the recovery module. Removing local recovery in the `w/o Local Recov.' variant leads to notable performance degradation—for example, a 1.38\% drop on HAR—highlighting the importance of explicitly recovering local temporal dependencies. Similarly, removing the entropy consistency constraint in the `w/o Local Consist.' variant also results in a performance decline, validating the necessity of encouraging consistent entropy across different contexts. In addition, further insight comes from the third dimension, which assesses the contribution of each individual context to temporal recovery. Among these variants, the `Complete' context proves most critical—its removal results in the largest performance drop, highlighting the essential role of capturing global temporal structures even when local temporal dependencies are recovered. Regarding the local contexts, the `Early' and `Late' contexts contribute more than the `Recovered' context, due to their potential overlap with the masked regions, thus providing partially recovered contexts to guide the recovery. Nevertheless, the `Recovered' context still remains crucial, as it directly targets the masked regions, encouraging alignment with the temporal dynamics of the source domain and further reinforcing the recovery process.

Lastly, we examine the effectiveness of the components within the enhancement module. Without anchor guidance, the `w/o Anchor Guid.' variant suffers from significant performance degradation. This is because the diversity maximization module, when used alone, may lead to negatively maximized diversity, which adversely affects the recovered domain and ultimately reduces performance, as discussed in Fig. \ref{fig:ARDM}. Introducing anchor guidance in the `w/o Anchor Guid. Aux.' variant improves performance; however, without the auxiliary objective to reinforce anchor guidance, its effectiveness remains limited. This highlights the necessity of incorporating the auxiliary component to strengthen the guidance. Furthermore, the `w/o Anchor Bank' variant, which generates anchors on a per-batch basis rather than maintaining a dedicated anchor bank, results in lower-quality anchors and diminished adaptation performance.


\subsection{Hyperparameter Sensitivity Analysis}
\label{sec:sensitivity}
We provide sensitivity analysis for the hyperparameters $\lambda_{LCA}$, $\lambda_{ARDM}$, the masking ratio, the anchor ratio, and the extraction proportion. Each hyperparameter is evaluated individually while keeping the others fixed to isolate its effect. For conciseness, results are reported using the average MF1 across all cross-domain scenarios.

\begin{figure}[ht]
    \begin{minipage}{0.5\textwidth}
        \centering
        \includegraphics[width=0.95\linewidth]{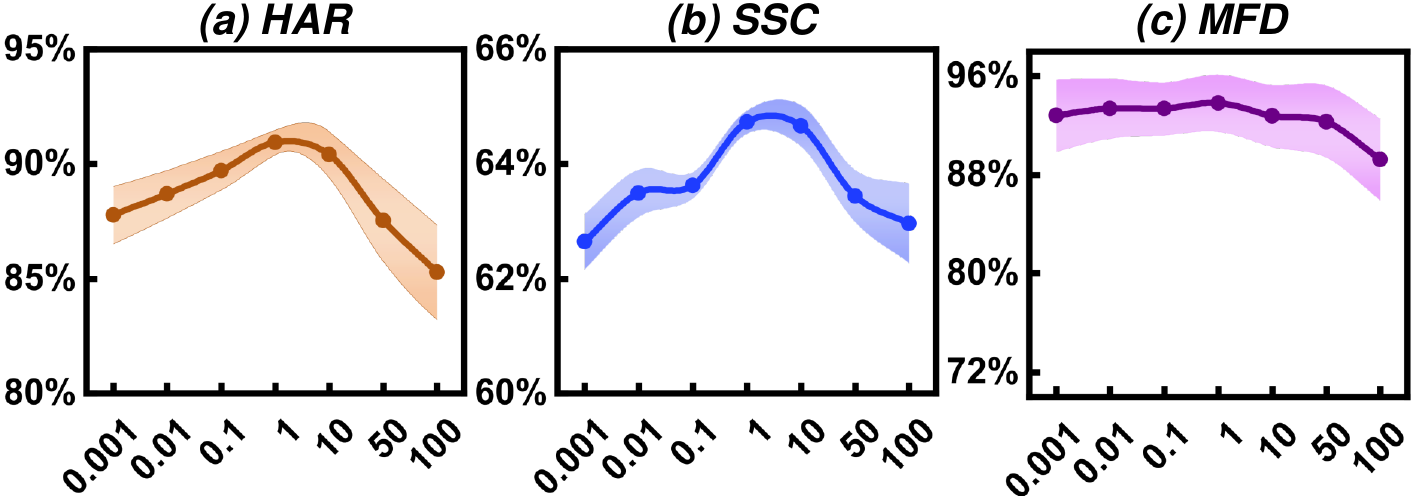}
        \vspace{-0.2cm}
        \caption{Analysis for $\lambda_{LCA}$.}
        \label{fig:LCA_sensivity}
    \end{minipage}%
    \hfill
    \begin{minipage}{0.5\textwidth}
        \centering
        \includegraphics[width=0.95\linewidth]{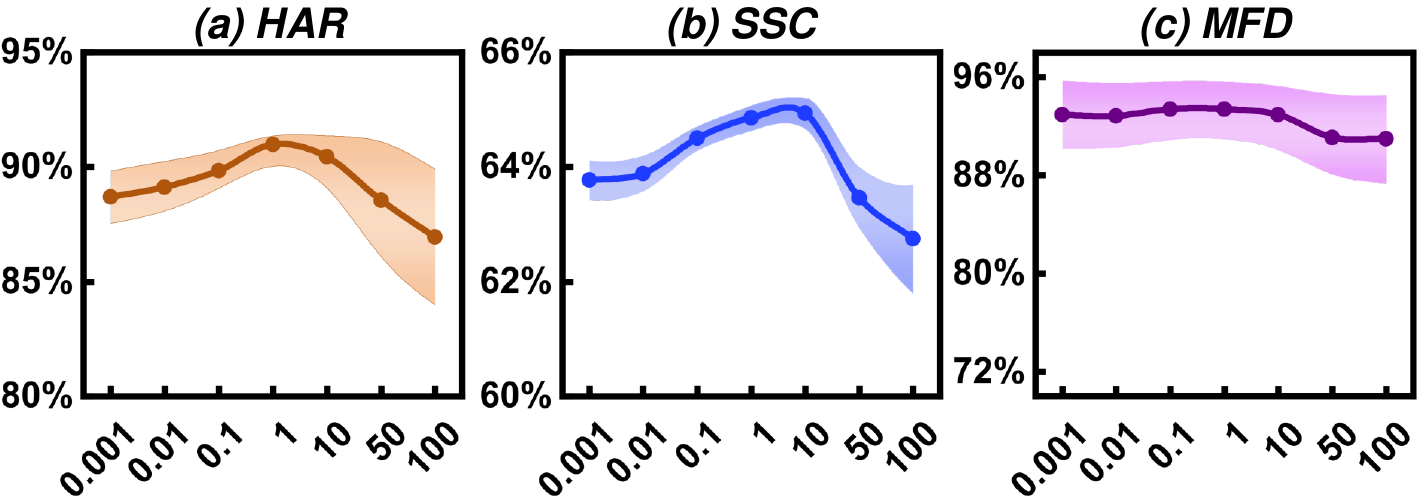}
        \vspace{-0.2cm}
        \caption{Analysis for $\lambda_{ARDM}$.}
        \label{fig:ARDM_sensivity}
    \end{minipage}
\end{figure}

\subsubsection{Effect of Hyperparameters $\lambda_{LCA}$ and $\lambda_{ARDM}$}
To fully evaluate the effects of $\lambda_{LCA}$ and $\lambda_{ARDM}$ on losses $\mathcal{L}_{LCA}$ and $\mathcal{L}_{ARDM}$, we adopted a wide range—[1e-3, 1e-2, 1e-1, 1, 10, 50, 100], with larger values indicating greater impacts. Fig. \ref{fig:LCA_sensivity} and \ref{fig:ARDM_sensivity} present the analysis for $\lambda_{LCA}$ and $\lambda_{ARDM}$, respectively. The results show that the performance of TemSR improves as $\lambda_{LCA}$ and $\lambda_{ARDM}$ increase, indicating that greater weights on these losses enhance performance, further highlighting their effectiveness. However, performance drops sharply when these values become too large, e.g., 50 or 100. For instance, with $\lambda_{LCA}=10\rightarrow100$, the performance on HAR decreases significantly, i.e., from around 91\% to 85\%. A similar trend is observed with $\lambda_{ARDM}$. These drops occur because, at higher values, the individual loss term dominates the adaptation process, overshadowing the contributions of other losses and thus negatively impacting adaptation. Meanwhile, excessive values also lead to instability, especially at 100. Based on these findings, the optimal range for both $\lambda_{LCA}$ and $\lambda_{ARDM}$ is between 1 and 10, offering a broad range to easily facilitate optimal performance for TemSR.


\subsubsection{Effect of Masking Ratio $p_m$}
\begin{figure}[htbp]
    \centering
    \includegraphics[width=.95\linewidth]{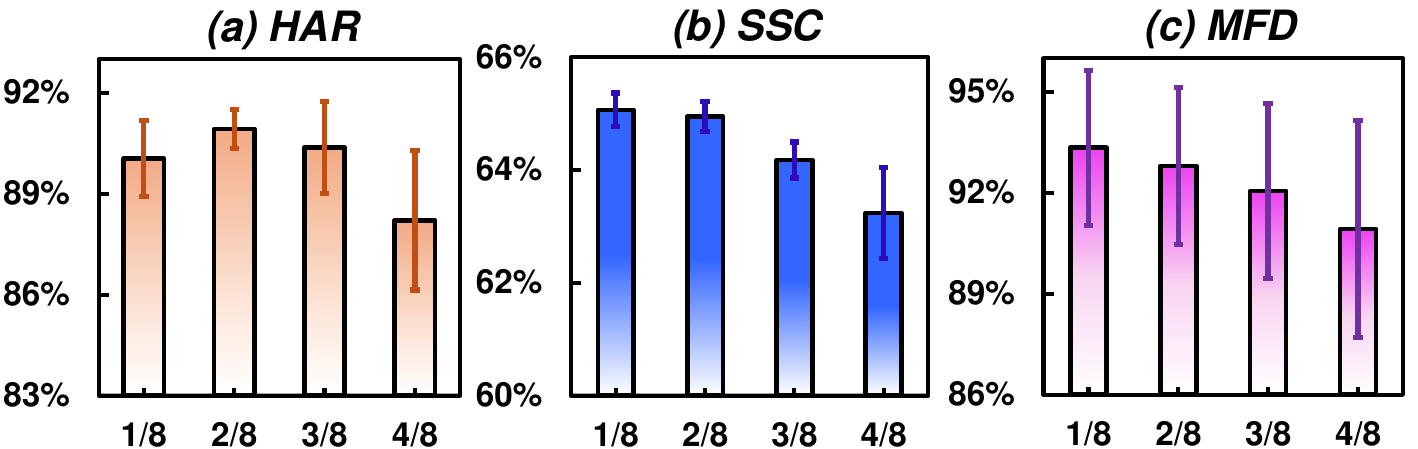}
    \caption{Analysis for Masking Ratio.}
    \label{fig:mask_sensivity}
\end{figure}
The masking ratio, which introduces diversity into the initial distribution to facilitate optimization toward a source-like distribution, is tested with values of [1/8, 2/8, 3/8, 4/8], with larger values indicating greater portions of the sample being masked. As shown in Fig. \ref{fig:mask_sensivity}, smaller masking ratios tend to yield better performance. As discussed in Sec. \ref{sec:enhancement}, although higher masking ratios introduce greater diversity into the recovered distribution, they also increase the risk of model collapse. In such cases, the recovery model may exploit shortcuts, such as recovering masked regions with constant values, leading to unstable performance. Conversely, while smaller masking ratios offer less inherent diversity, our proposed recovery diversity maximization loss mitigates this limitation by encouraging sufficient diversity while maintaining its fidelity to the source domain. Therefore, smaller masking ratios—such as 1/8 or 2/8—strike a favorable balance between diversity and stability, and are recommended for achieving optimal adaptation performance.

\subsubsection{Effect of Anchor Ratio}
\begin{figure}[htbp]
    \centering
    \includegraphics[width=.95\linewidth]{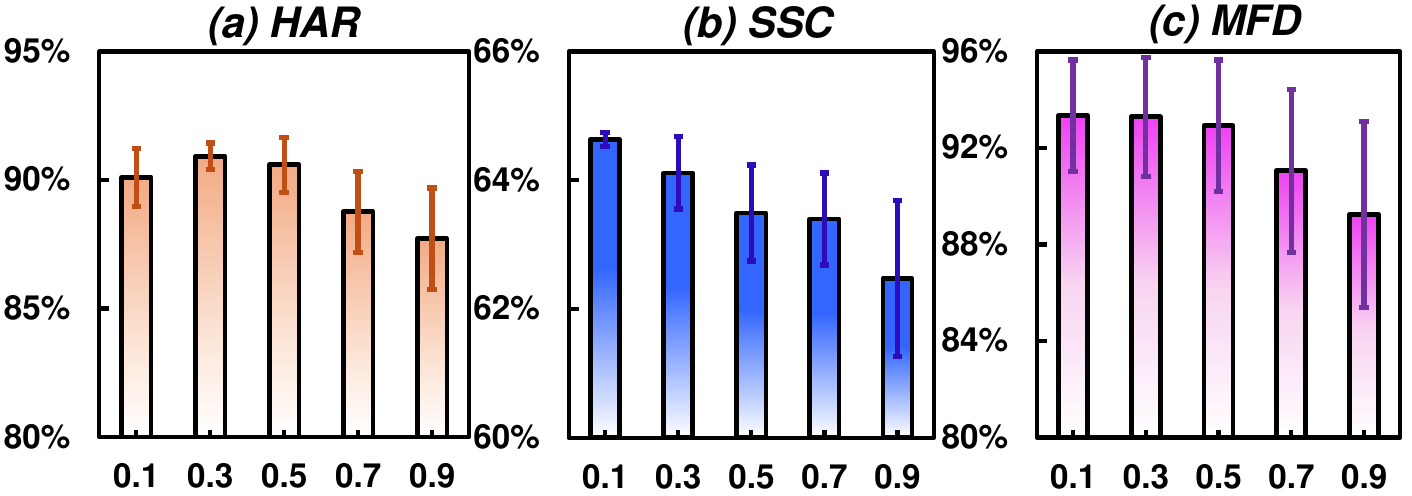}
    \caption{Analysis for Anchor Ratio.}
    \label{fig:anchor_sensivity}
\end{figure}
The anchor ratio, which determines the $k$ samples with the lowest entropy to generate the representative anchor, is evaluated using [0.1, 0.3, 0.5, 0.7, 0.9], with larger values indicating more samples selected for generating the anchor sample. For example, 0.1 represents the 10\% of samples with lowest entropy being selected for anchor generation. Fig. \ref{fig:anchor_sensivity} shows the sensitivity of TemSR to different anchor ratios, where smaller anchor ratios tend to yield better results. This is because samples with the lowest entropy are more likely to produce higher-quality anchors with greater confidence. In contrast, larger anchor ratios may include samples with lower confidence (with larger entropy), leading to less accurate anchors and, consequently, poorer guidance during the adaptation process. From these results, anchor ratios of 0.1 or 0.3 are recommended for generating effective anchors to enhance performance. 
 
\subsubsection{Effect of Extraction Proportion $p_s$}
\begin{figure}[htbp]
    \centering
    \includegraphics[width=.95\linewidth]{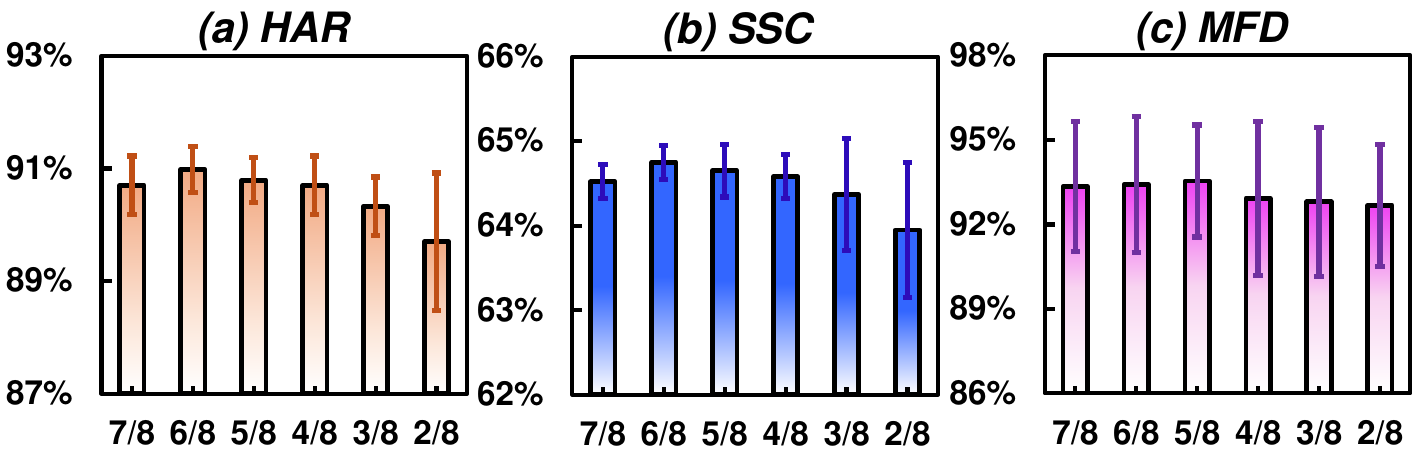}
    \caption{Analysis for extraction proportion.}
    \label{fig:LCA_propor_sensivity}
\end{figure}
The extraction proportion determines the amount of local information preserved in each local context. To evaluate its impact, we tested values within [7/8,6/8,5/8,4/8,3/8,2/8]. A value of 1 corresponds to the full sequence, reflecting purely global context, while smaller values indicate that shorter ranges are extracted, focusing on local temporal recovery within narrower regions. From the analysis in Fig. \ref{fig:LCA_propor_sensivity}, we observe that reducing the extraction proportion—for example, from 7/8 to 6/8—can improve performance. This improvement arises because a lower proportion restricts entropy minimization to a narrower temporal range, aligning the local distribution of recovered samples with the source distribution and thereby enhancing the recovery of pattern transitions in local regions. However, when the proportion becomes too small, such as 2/8, each local context contains limited information from the recovered sample, making it difficult to capture meaningful local temporal dependencies. As a result, the recovery model may misinterpret the entropy minimization objective and generate ineffective source-like distributions, ultimately degrading adaptation performance. Thus, extraction proportions of 6/8 or 5/8 offer a better trade-off for optimal local temporal recovery.
\subsection{Discussions}
\subsubsection{Distribution Discrepancy Changes}
\begin{figure}[htbp]
    \centering
    \includegraphics[width=0.9\linewidth]{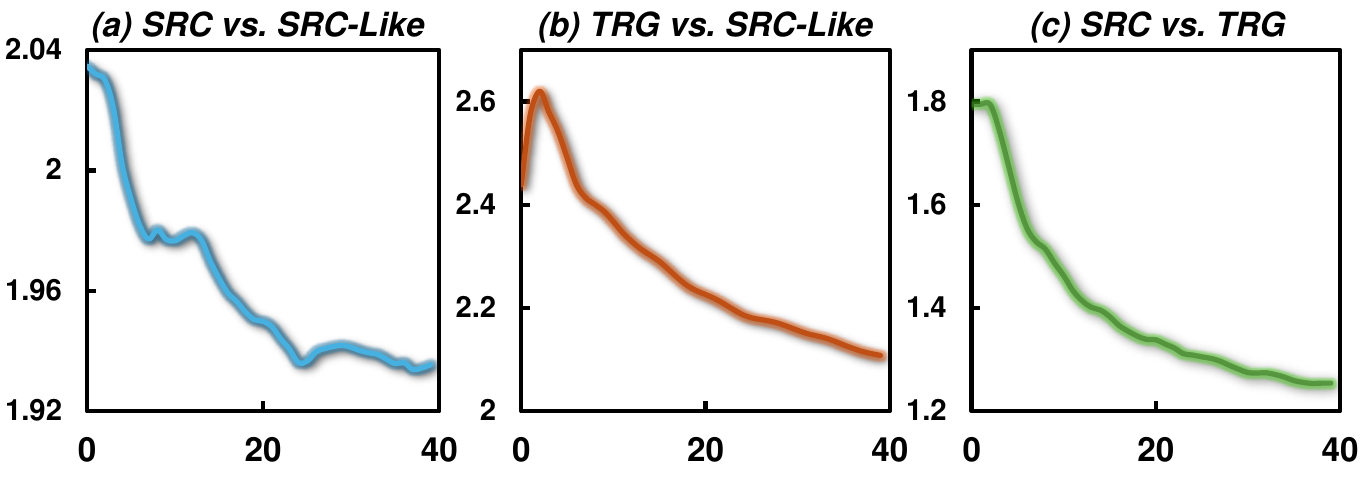}
    \caption{Distribution discrepancies changes (Original source domain used only for computing discrepancy without directly involved in adaptation).}
    \label{fig:distribution_changing}
\end{figure}
The core objective of TemSR is to recover a source-like domain and then transfer knowledge from this recovered domain to the target domain. Achieving this requires ensuring that the recovered source-like distribution closely resembles the original source distribution and that the domain discrepancy between the source-like and target domains is minimized. By doing so, TemSR can effectively reduce the gap between the original source domain and the target domain. To evaluate this process intuitively, we visualize the evolution of distribution discrepancies among the source (SRC), SRC-like, and target (TRG) domains throughout the adaptation stage, as shown in Fig. \ref{fig:distribution_changing}. The discrepancies are quantified using KL divergence, a widely used metric for measuring distributional differences \cite{zhang2024properties}. Notably, in this analysis, the source distribution is used only for calculating discrepancies and is not directly involved for adaptation. 

From Fig. \ref{fig:distribution_changing}, we observe that the discrepancy between the source and source-like domains decreases steadily during the adaptation stage, indicating that the recovered source-like distribution increasingly resembles the original source distribution. In the early epochs, prior to alignment, the domain gap between the target and source-like distributions initially increases. However, once alignment begins, this gap gradually narrows. Additionally, the overall discrepancy between the source and target domains decreases over time. By the end of the adaptation stage, the source and target domains can be well aligned, demonstrating TemSR’s effectiveness in bridging domains without requiring access to the source data.

\subsubsection{Comparisons with Foundation Model}
\begin{table*}[htbp]
  \centering
  \caption{Detailed results of the ten HAR, SSC, and MFD cross-domain scenarios in terms of MF1 score comparing with TSFM.}
  \begin{adjustbox}{width = .95\linewidth,center}
    \begin{tabular}{clccccccccccc}
    \toprule
    \toprule
    \multirow{4}[4]{*}{HAR} & Models & 2$\rightarrow$11  & 12$\rightarrow$16 & 9$\rightarrow$18  & 6$\rightarrow$23  & 7$\rightarrow$13  & 18$\rightarrow$27 & 20$\rightarrow$5  & 24$\rightarrow$8  & 28$\rightarrow$27 & 30$\rightarrow$20  & Avg. \\
\cmidrule{2-13}          & SRC & 95.69$\pm$5.72  & 67.13$\pm$9.83  & 70.07$\pm$4.71  & 81.01$\pm$14.9  & 84.5$\pm$12.08  & 85.95$\pm$5.00  & 63.30$\pm$4.13  & 71.59$\pm$8.56  & 50.24$\pm$5.92  & 67.91$\pm$9.21  & 73.73$\pm$2.68 \\
          & TSFM    & 75.34$\pm$1.19 & \textbf{80.92$\pm$0.34} & 70.14$\pm$2.06 & 71.58$\pm$2.77 & 89.85$\pm$0.24 & 98.87$\pm$0.00 & 68.26$\pm$0.33 & 88.20$\pm$1.08 & 87.96$\pm$2.69 & 82.72$\pm$4.63 & 81.38$\pm$0.04 \\
          & TemSR  & \textbf{100.0$\pm$0.00} & 64.21$\pm$3.04 & \textbf{93.65$\pm$2.02} & \textbf{97.82$\pm$1.89} & \textbf{98.95$\pm$0.01} & \textbf{100.0$\pm$0.00} & \textbf{82.32$\pm$0.73} & \textbf{100.0$\pm$0.00} & \textbf{100.0$\pm$0.00} & \textbf{84.10$\pm$5.52} & \textbf{92.10$\pm$0.33} \\
        
    \midrule
    \midrule
    \multirow{4}[4]{*}{SSC} & Models & 16$\rightarrow$1  & 9$\rightarrow$14  & 12$\rightarrow$5  & 7$\rightarrow$18  & 0$\rightarrow$11  & 3$\rightarrow$19  & 18$\rightarrow$12  & 13$\rightarrow$17  & 5$\rightarrow$15  & 6$\rightarrow$2   & Avg. \\
\cmidrule{2-13}          & SRC & 52.93$\pm$3.42  & 63.99$\pm$8.04  & 48.79$\pm$3.31  & 62.33$\pm$3.86  & \textbf{50.43$\pm$6.26}  & 47.38$\pm$3.36  & 38.35$\pm$2.03  & 43.80$\pm$0.12  & 60.13$\pm$6.36  & 55.67$\pm$2.20  & 52.38$\pm$0.47 \\
          & TSFM    & \textbf{68.15$\pm$0.07} & 60.08$\pm$1.17 & 65.32$\pm$3.11 & 67.59$\pm$0.37 & 47.73$\pm$1.32 & 50.36$\pm$0.35 & 54.67$\pm$0.34 & 44.57$\pm$1.27 & 52.65$\pm$2.58 & 63.29$\pm$2.33 & 57.44$\pm$0.43 \\
          & TemSR  & 62.51$\pm$1.09  & \textbf{72.60$\pm$0.74}  & \textbf{66.70$\pm$1.91}  & \textbf{72.15$\pm$1.01}  & 49.62$\pm$1.88  & \textbf{65.87$\pm$0.53}  & \textbf{60.32$\pm$0.97}  & \textbf{57.56$\pm$2.07}  & \textbf{66.50$\pm$2.07}  & \textbf{64.82$\pm$1.78}  & \textbf{63.86$\pm$0.58} \\
    \midrule
    \midrule
    \multirow{4}[4]{*}{MFD} & Models & 0$\rightarrow$1   & 1$\rightarrow$0   & 1$\rightarrow$2   & 2$\rightarrow$3   & 3$\rightarrow$1   & 0$\rightarrow$3   & 1$\rightarrow$3   & 2$\rightarrow$1   & 3$\rightarrow$0   & 3$\rightarrow$2   & Avg. \\
\cmidrule{2-13}          & SRC & 26.26$\pm$5.04  & 68.63$\pm$6.22  & 72.66$\pm$0.95  & \textbf{96.90$\pm$1.38}  & 99.02$\pm$1.07  & 42.13$\pm$8.06  & 96.25$\pm$3.72  & 86.96$\pm$0.58  & 46.42$\pm$2.42  & 71.71$\pm$6.54  & 70.69$\pm$2.61 \\
          & TSFM    & 29.77$\pm$5.56 & 73.10$\pm$2.12 & \textbf{85.61$\pm$0.06} & 93.89$\pm$1.15 & 98.20$\pm$0.23 & 55.83$\pm$8.69 & 96.90$\pm$0.23 & 92.09$\pm$0.06 & 56.43$\pm$1.40 & \textbf{88.25$\pm$0.03} & 77.01$\pm$0.31 \\
          & TemSR  & \textbf{99.97$\pm$0.05}  & \textbf{87.03$\pm$4.05}  & 84.47$\pm$5.88  & 95.23$\pm$3.85  & \textbf{100.0$\pm$0.00}  & \textbf{99.95$\pm$0.05}  & \textbf{100.0$\pm$0.00}  & \textbf{96.67$\pm$4.21}  & \textbf{87.17$\pm$1.56}  & 81.96$\pm$5.09  & \textbf{93.24$\pm$1.83} \\
    \bottomrule
    \bottomrule
    \end{tabular}%
    \end{adjustbox}
  \label{tab:comp_tsfm}%
\end{table*}%

In the era of Large Models (LMs), foundation models have gained popularity across various domains \cite{chang2024survey,liang2024foundation}. Pretrained on extensive datasets, LMs typically exhibit strong generalizability to new domains, making them promising candidates for addressing SFUDA challenges. To investigate this potential and provide a comprehensive comparison with our TemSR in the context of TS-SFUDA, we employed Time-Series Foundation Models (TSFMs), which are pretrained on large-scale TS data and demonstrate strong generalization on TS tasks. 

Specifically, we evaluated TSFMs using Mantis \cite{feofanov2025mantis}, an open-source foundation model designed for TS classification. As benchmarks such as HAR and SSC also involve classification tasks, comparing with Mantis ensures a fair comparison. The performance comparisons are presented in Table~\ref{tab:comp_tsfm}, where TSFM outperforms SRC (the source only baseline without adaptation), highlighting its generalizability and potential for SFUDA. However, as TSFMs are pretrained to capture general patterns across TS data, they struggle to model task-specific dependencies tailored to a particular domain. Moreover, TSFMs are not explicitly designed to transfer temporal dependencies across domains, limiting their effectiveness in TS-SFUDA settings. As a result, their performance remains significantly lower than our TemSR, which achieves substantial improvements, e.g., outperforming TSFM by 10.72\% on HAR.

Beyond performance, we also compare the complexity of both approaches. TSFMs typically require more parameters due to their design for general-purpose representation learning; for instance, Mantis has 8 million parameters. In contrast, TemSR adopts a lightweight CNN backbone with only 200.1K parameters, making it more practical. Moreover, TSFMs still require additional fine-tuning to align with the target task’s label space. In our experiments, Mantis required 324.8 seconds to fine-tune for a single cross-domain scenario, whereas TemSR completed adaptation for ten scenarios in just 83.24 seconds.

These comparisons suggest that while foundation models offer promising generalization capabilities, specialized approaches like TemSR still remain essential for achieving superior and efficient performance in TS-SFUDA settings.



\subsubsection{Computation Complexity Analysis}
Model complexity analysis is crucial for assessing the practicality of TS-SFUDA techniques in real-world scenarios. As all TS-SFUDA baselines utilize the same backbone, standard complexity metrics such as FLOPs or parameter counts offer limited insights. Instead, we assess computational complexity by measuring the actual runtime, as each method incorporates distinct operations that directly affect pretraining and adaptation time. Specifically, we run each method once across ten cross-domain scenarios on HAR with an RTX 3080Ti GPU. To ensure fairness, the comparison is limited to SFUDA techniques only. 

From the results in Table \ref{tab:runningtime}, we observe that traditional methods generally require less time, as they lack the operations for transferring source temporal dependencies—though this also results in weaker performance. While MAPU and TemSR incorporate additional operations, the extra runtime required is limited, adding only a few seconds to total runtime. Notably, compared to MAPU, TemSR does not rely on specific pretraining steps, thus resulting in reduced overall runtime. This demonstrates that TemSR not only effectively recovers temporal dependencies during the adaptation stage but also achieves this with limited computational resources, ensuring practical applicability.

\begin{table}[htbp]
  \centering
  \caption{Running time comparisons of TS-SFUDA techniques.}
  \begin{adjustbox}{width = 1\linewidth,center}
    \begin{tabular}{ccccccccc}
    \toprule
    \toprule
    Models & SHOT  & NRC   & AaD   & BAIT & SF(DA)$^2$ & MAPU & TFDA & TemSR \\
    \midrule
    Training/s & 70.43 & 66.63 & 72.94 & 74.36 & 95.03 & 89.79 & 79.86 & 83.24 \\
    \bottomrule
    \bottomrule
    \end{tabular}%
    \end{adjustbox}
  \label{tab:runningtime}%
\end{table}%
\section{Conclusion}
To transfer temporal dependencies across domains without source data or source-specific pretraining, we propose Temporal Source Recovery (TemSR), an effective and practical framework for TS-SFUDA. TemSR features a masking-recovery-optimization process that exploits the intrinsic properties of TS data to generate a source-like domain with recovered temporal dependencies, thereby enabling their effective transfer to the target domain with standard UDA techniques. To further refine the source-like domain, we enhance the optimization as local context-aware regularization to restore local temporal dependencies and meanwhile introduce an anchor-based recovery diversity maximization loss to promote diversity within the source-like distribution. Additional analysis of distribution discrepancy changes between source, source-like, and target domains confirms TemSR's ability to recover the source-like domain, ultimately reducing gaps between the source and target domains. Extensive experiments further demonstrate the SOTA performance of TemSR, which even surpasses the existing TS-SFUDA method that relies on source-specific designs.

\bibliography{refer}
\bibliographystyle{IEEEtran}

\end{document}